\pdfoutput=1

\documentclass[11pt]{article}

\usepackage[]{ACL2023}

\usepackage{times}
\usepackage{latexsym}
\usepackage{amsmath,amsfonts,amssymb}

\usepackage[T1]{fontenc}

\usepackage[utf8]{inputenc}

\usepackage{microtype}
\usepackage{inconsolata}

\usepackage[utf8]{inputenc}
\usepackage[T1]{fontenc}
\usepackage{hyperref}
\usepackage{url}
\usepackage{graphicx}
\usepackage{caption}
\usepackage{subcaption}
\usepackage{booktabs}
\usepackage{multirow}
\usepackage{xspace}
\usepackage{tabularx}
\usepackage{array}
\usepackage{makecell}
\usepackage{enumitem}

\usepackage{cleveref}
\crefname{figure}{Fig.}{Figs.}
\crefname{table}{Table}{Tables}
\crefname{appendix}{App.}{Apps.}
\crefname{section}{\S}{\S\S}
\Crefname{section}{\S}{\S\S}
\crefname{equation}{Eq.}{Eqs.}
\crefname{algorithm}{Alg.}{Algs.}
\crefname{algocf}{Alg.}{Algs.}

\usepackage[ruled,linesnumbered]{algorithm2e}
\usepackage[noend]{algpseudocode}  
\algrenewcommand\algorithmicindent{1.0em}%

\newcommand{\rightcomment}[1]{{\color{gray} \(\triangleright\){\footnotesize\textit{#1}}}}
\algrenewcommand{\algorithmiccomment}[1]{\hfill \rightcomment{#1}}  
\algnewcommand{\LineComment}[1]{\State\rightcomment{#1}}
\algnewcommand{\LinesComment}[1]{\State\rightcomment{\parbox[t]{.95\linewidth-\leftmargin-\widthof{\(\triangleright\) }}{#1}}}

\algrenewcommand\alglinenumber[1]{{\tiny\color{black!50}#1.}\hspace{-2pt}}
\newcommand{\algorithmicfunc}[1]{\textbf{def} {#1}:}
\algdef{SE}[FUNC]{Func} {EndFunc}[1]{\algorithmicfunc{#1}}{}
\makeatletter
\ifthenelse{\equal{\ALG@noend}{t}}%
  {\algtext*{EndFunc}}
  {}%

\newcommand{\formalismname}{\textsc{MathWorld}\xspace} 
\newcommand{\msp}{MSP\xspace} 
\newcommand{\msps}{MSPs\xspace} 
\newcommand{\llm}{LLM\xspace} 
\newcommand{\llms}{LLMs\xspace}
\newcommand{\aproblem}{\mathbf{s}}
\newcommand{\body}{\mathbf{b}}

\newcommand{\problemelem}[1]{s_{#1}}
\newcommand{\question}{q}
\newcommand{\aworldmodel}{g}
\newcommand{\asubwm}[1]{g_{#1}}
\newcommand{\refexpr}{r}
\newcommand{\lform}[1]{m_{#1}}
\newcommand{\defn}[1]{\textbf{#1}}
\newcommand{\Land}{\hspace{0.1em}\land\hspace{0.1em}}
\newcommand{\transfer}{\textsc{Transfer}\xspace}
\newcommand{\rate}{\textsc{Rate}\xspace}
\newcommand{\explicit}{\textsc{Comparison}\xspace}
\newcommand{\comparison}{\textsc{Comparison}\xspace}
\newcommand{\partwhole}{\textsc{PartWhole}\xspace}
\newcommand{\mawps}{\textsc{MAWPS}\xspace}
\newcommand{\asdiv}{\textsc{ASDiv-A}\xspace}
\newcommand{\svamp}{\textsc{SVAMP}\xspace}
\usepackage[group-separator={,},group-minimum-digits={3}]{siunitx}

\newcommand{\saveforcameraready}[1]{#1}

\usepackage{todonotes}
\makeatletter
\newcommand*\iftodonotes{\if@todonotes@disabled\expandafter\@secondoftwo\else\expandafter\@firstoftwo\fi}  
\makeatother

\newcommand{\ethz}{\text{\normalfont $\otimes$}}
\newcommand{\cls}{\text{\normalfont $\pm$}}
\newcommand{\nyu}{\text{\normalfont $\div$}}

\title{World Models for Math Story Problems
}

\author{
Andreas Opedal$^{\ethz,\cls}$~\;~
Niklas Stoehr$^{\ethz}$~\;~
Abulhair Saparov$^{\nyu}$~\;~
Mrinmaya Sachan$^{\ethz}$
\\
$^{\ethz}$ETH Z{\"u}rich \qquad $^{\nyu}$New York University \\
$^{\cls}$Max Planck ETH Center for Learning Systems
\\
\footnotesize
\href{mailto:andreas.opedal@inf.ethz.ch}{\texttt{andreas.opedal@inf.ethz.ch}}~\;~
\href{mailto:niklas.stoehr@inf.ethz.ch}{\texttt{niklas.stoehr@inf.ethz.ch}}~\;~
\href{mailto:as17582@nyu.edu}{\texttt{as17582@nyu.edu}}~\;~
\href{mailto:mrinmaya.sachan@inf.ethz.ch}{\texttt{mrinmaya.sachan@inf.ethz.ch}}~\;~
}

\begin{document}
\maketitle
\begin{abstract}
Solving math story problems is a complex task for students and NLP models alike, requiring them to understand the world as described in the story and reason over it to compute an answer. Recent years have seen impressive performance on automatically solving these problems with large pre-trained language models and innovative techniques to prompt them. However, it remains unclear if these models possess accurate representations of mathematical concepts. This leads to lack of interpretability and trustworthiness which impedes their usefulness in various applications. In this paper, we consolidate previous work on categorizing and representing math story problems and develop $\formalismname$, which is a graph-based semantic formalism specific for the domain of math story problems. With $\formalismname$, we can assign world models to math story problems which represent the situations and actions introduced in the text and their mathematical relationships. We combine math story problems from several existing datasets and annotate a corpus of $1,019$ problems and $3,204$ logical forms with $\formalismname$. Using this data, we demonstrate the following use cases of $\formalismname$: (1) prompting language models with synthetically generated question-answer pairs to probe their reasoning and world modeling abilities, and (2) generating new problems by using the world models as a design space.
\newline
\newline
\vspace{1.5em} 
\hspace{.5em}\includegraphics[width=1.25em,height=1.25em]{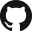}{\hspace{.75em}\parbox{\dimexpr\linewidth-2\fboxsep-2\fboxrule}{\url{https://github.com/eth-nlped/mathworld}}}
\vspace{-2em}
\end{abstract}

\section{Introduction}
\begin{figure}[t]
     \centering
     \includegraphics[width=1.00\linewidth]{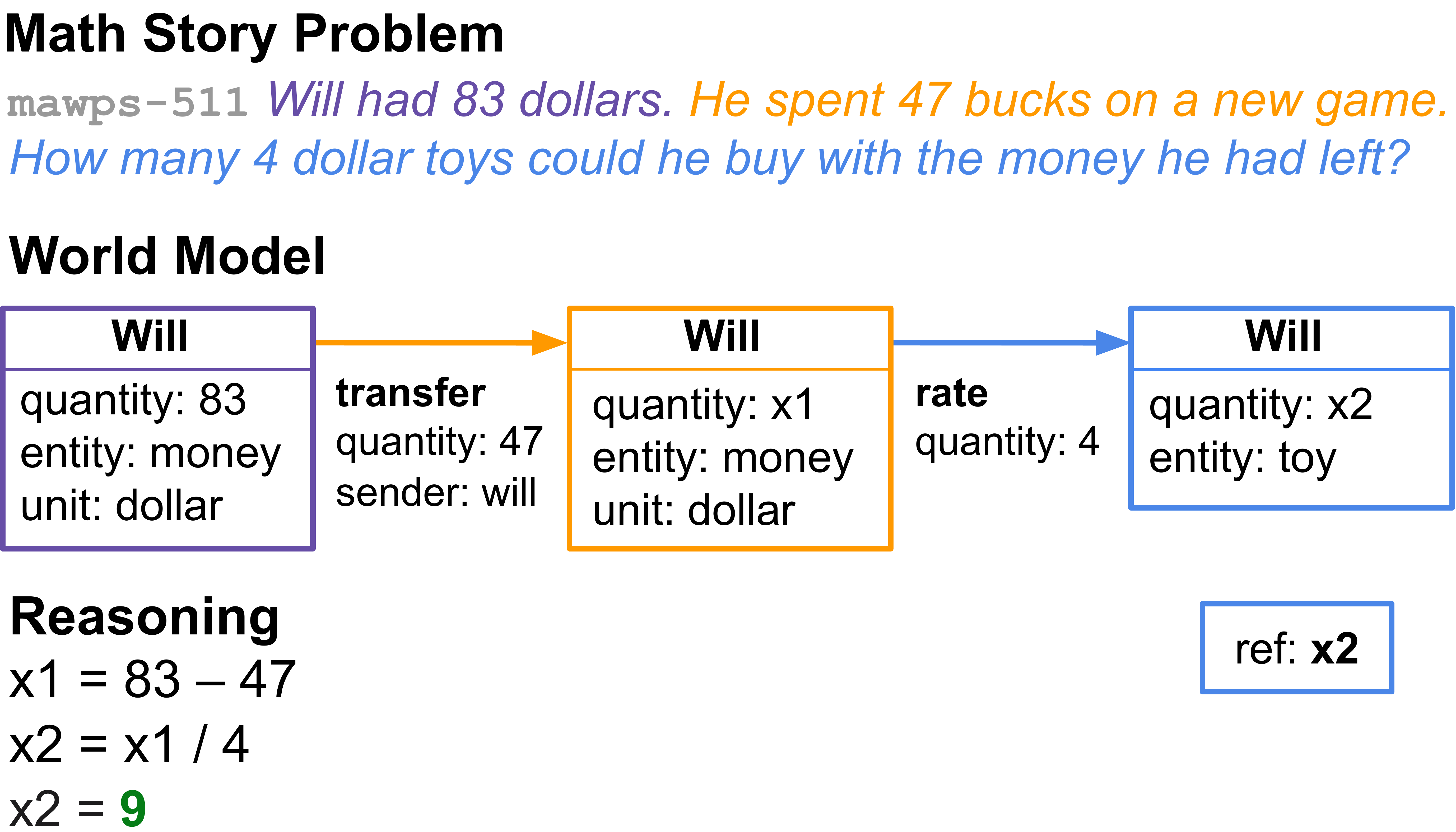}
     \caption{An example of a world model in \formalismname. 
     \formalismname can be used to develop interpretable \msp solvers, to study the reasoning of \llms and as a design space for generation of new \msps. 
     }
     \label{fig:overview}
     \vspace{-0.3cm}
\end{figure}

Math story problems (\msps) are short narrative texts that describe a dynamic situation in the world consisting of entities, actions and states,
followed by a quantitative question about the world, as displayed in \cref{fig:overview}.
The task of automatically solving \msps has received much research attention in NLP.
While earlier models for solving \msps \citep{hosseini-etal-2014-learning,kushman-etal-2014-learning,roy-roth-2015-solving} 
focused on extracting various features from text to learn probabilistic models,
recent efforts have used pre-trained large language models (\llms) \citep[\emph{inter alia}]{yang_nt5_2021, Drori22, Lewkowycz2022}. Although they display high performance on benchmarks,
it has been shown that such neural models tend to rely heavily on shallow heuristics \citep{patel-etal-2021-nlp,stolfo-etal-2022-causal}, raising questions about whether the models can indeed ``understand'' \msps and robustly solve them.\looseness=-1

From the human side, solving \msps requires a wide set of skills.
A student
must not only perform a set 
of given computations, but first be able to process the text and map it into a corresponding world model 
that represents the situation described in text \citep{CUMMINS1988405, stern_1993}.
Inspired by this, we take a step towards 
developing more interpretable solvers 
and
introduce $\formalismname$, a semantic world model framework for \msps.

$\formalismname$ can be viewed as a formalism for reasoning in dynamical problem settings 
\citep{McCarthy1963SituationsAA, reiter-frame-1991}, specific to the domain of \msps.
It represents each problem as a directed graph called a \emph{world model} (\cref{sec:world-model}). The nodes in a world model are containers (\cref{sec:containers}) representing entities' possession of some quantity \citep{hosseini-etal-2014-learning} and the edges represent 
various types of mathematical relations between the quantities (\cref{sec:relations}). The relations correspond to mathematical concepts that have been previously shown to cover a vast majority of \msps \citep{mitra-baral-2016-learning, roy-declarative}.
We annotate a \formalismname dataset consisting of $1,019$ English
\msps 
from various widely-used datasets \citep{koncel-kedziorski-etal-2016-mawps, miao-etal-2020-diverse, patel-etal-2021-nlp}, which we make publicly available.
\looseness=-1 

There are several potential use cases of \formalismname, of which we discuss three. 
First, one natural application is that of developing interpretable \msp solvers. A solver using \formalismname follows two steps: (i) semantic parsing and (ii) reasoning. The semantic parser takes an \msp text and outputs a world model based on the explicit information in the text. The reasoner then takes the world model 
and solves the problem based on the quantities and their relations. 
Our experiments show that \llms struggle to build accurate and well-formed world models; we encourage future work to develop stronger semantic parsers for \formalismname.\looseness=-1

\begin{table*}[]
\small
\centering
\renewcommand{\arraystretch}{2.35} 
\setlength{\tabcolsep}{0.8em} 
\begin{tabular}{l|c|c|c|c|c}
 & \makecell{\textbf{Arithmetic} \\ \textbf{coverage}} & \makecell{\textbf{Conceptual} \\ \textbf{coverage}} & \makecell{\textbf{Semantic} \\ \textbf{granularity}} & \textbf{Annotations?} & \makecell{\textbf{Mapping to} \\ \textbf{formal logic?}} \\\toprule
 \formalismname & $(+,-,\times,\div)$ & \makecell{Transfer \\ Rate \\Comparison \\ Part-whole} & World model & Yes & Yes \\\hline
 \citet{hosseini-etal-2014-learning} & $(+,-)$ & Transfer & World model & No & No \\\hline
 \citet{mitra-baral-2016-learning} & $(+,-)$ & \makecell{Transfer \\Comparison (add) \\ Part-whole} & \makecell{Concepts \\ \& equations} & Yes & No \\\hline
 \citet{roy-declarative} & $(+,-,\times,\div)$ & \makecell{Transfer \\ Rate \\Comparison \\ Part-whole} & \makecell{Concepts \\ \& equations} & No & No \\\bottomrule
\end{tabular}
\caption{Comparison between \formalismname and other \msp works that use a more fine-grained symbolics than equations alone. ``Annotations'' refers to whether those symbolics are explicitly provided in the dataset.}
\label{table:related-work}
\end{table*}

Another use case of \formalismname is as a tool to study the reasoning capabilities of existing solvers. 
For instance, we can use the world model annotations to automatically generate synthetic subquestions for the \msps. Using such subquestions, we give empirical evidence that 
GPT-3 \citep{brown-gpt-3} benefits from the structured knowledge derived by world models in its ability to solve \msps.
We further use our synthetic questions to understand if GPT-3 can indeed answer these intermediate questions about the world described in the \msps, and not just the final question. 
We find that for problems where GPT-3 answers the final question correctly, it can only answer 64\% of the intermediate questions. This 
suggests that GPT-3 is not accurately building world models for these problems but might be relying on reasoning shortcuts.\looseness=-1

Finally, \formalismname\ can be considered as a design space for generating interesting new \msps.
We illustrate the usefulness of \formalismname for the task of generating \msps\ by prompting an \llm using the world model annotations.

\section{Related Work}

\paragraph{Math story problems in NLP.}
Although the problem of automatically solving \msps has gathered substantial interest in NLP
\citep{roy-roth-2015-solving, kushman-etal-2014-learning, huang-etal-2017-learning, amini-etal-2019-mathqa, xie-goal-driven, Drori22}, the focus has traditionally been on improving answer accuracy rather than providing didactic human-interpretable solutions \citep{shridhar-etal-2022-automatic}. Some approaches map the text to expression trees \citep{koncel-kedziorski-etal-2015-parsing, yang-logic-solver-22, roy_unit_dependency_17} or explicitly model arithmetic concepts \citep{mitra-baral-2016-learning, roy-declarative}. However, few if any computational works have attempted to solve \msps by using mental models \citep{johnson-laird-1983}, which is a common framework for analyzing how humans solve \msps \citep{Kintsch1985-KINUAS}. 
Taking inspiration from mental models of \msps, we offer \formalismname as a computational model (fully expressible in first-order logic, \cref{sec:app-fol}) which represents reasoning steps, arithmetic concepts and fictional elements in a human-readable graph format. We hope that such an approach can support intelligent tutoring systems \citep{anderson-cognitive-1995}, e.g., by delivering feedback and hints \citep{zhou-hints-1999, fossati-2008-role} or generating new \msps \citep{Polozov2015PersonalizedMW, koncel-kedziorski-etal-2016-theme, srivastava-goodman-2021-question}.\looseness=-1

In particular, we draw inspiration from \citet{hosseini-etal-2014-learning}, who propose a symbolic approach that maps the text to container-based states. 
However, their symbolic representation is purely extracted from syntactic rules without human annotation. Further, their approach only covers problems that involve a transfer of some quantity between some actors (although they do not use that terminology), requiring addition and/or subtraction. In contrast, $\formalismname$ is more closely tied to the \msp's semantics. It covers a strictly larger set of problem types, involving more concepts and all four basic arithmetic operators $(+,-,\times,\div)$. See \cref{table:related-work} for a comparison between \formalismname and \citet{hosseini-etal-2014-learning}, as well as \citet{mitra-baral-2016-learning} and \citet{roy-declarative} from which we adopt the taxonomy over arithmetic concepts.

\paragraph{Reasoning with large language models.} 
\llms have displayed impressive performance on numerical reasoning tasks \citep{brown-gpt-3, chowdhery-palm}, particularly by the help of careful prompt engineering \citep{wei_chain_of_thought, shridhar2023distilling, zhou-22-least}.
While language models have been argued to be intrinsically limited in their ability to perform human-like reasoning \citep{bender-koller-2020-climbing}, 
the mechanism by which they find answers in complex reasoning tasks is currently an active area of research \citep{tafjord-etal-2021-proofwriter, SaparovHe22}. \formalismname provides ground truth world model annotations, which is valuable in such studies (as demonstrated in \cref{sec:probing}).
One other aspect of \llms that may limit them when applied to reasoning is that they produce natural language text, which may be ambiguous and diverse. 
These considerations motivate us to study \msps as
structured representations of meaning, which can in turn be used to generate natural language \citep{saparov-towards-22}.

\paragraph{Semantic parsing.}
$\formalismname$ can be viewed as a domain-specific semantic formalism. Our work thus also relates closely to semantic parsing,
particularly of graph-based structures \citep{banarescu-etal-2013-abstract,cai-lam-2019-core,zhang-etal-2019-amr,bai-etal-2022-cross}.
However, while most other formalisms consider meaning only at the sentence level,
our world model graphs span the meaning across multiple sentences.

\section{$\formalismname$}\label{sec:world-model}

In this section, we present our world model formalism $\formalismname$. We formalize an \msp as a sequence of $n$ sentences $\aproblem=\problemelem{1} \circ \dots \circ \problemelem{n}$. It can be separated into a \defn{body} $\body$ and a \defn{question} $\question$, such that $\aproblem = \body \circ \question$. The body is further partitioned into a sequence of $n-1$ declarative sentences $\body =\problemelem{1}\circ \dots \circ \problemelem{n-1}$ and the question consists of a single interrogative sentence $\question = \problemelem{n}$. \looseness=-1

World models in $\formalismname$ are directed and labelled graphs, denoted $\aworldmodel$.\footnote{The graphs may be cyclic. Although in practice, they tend to be acyclic. 
} We refer to the nodes of the graph as \defn{containers} (\cref{sec:containers}) and the edges of the graph as \defn{relations} (\cref{sec:relations}). Each container and relation is labelled with a set of properties. One such property is the \defn{quantity}, which may be either an explicit number mentioned in text or a variable representing an unknown number.
The containers and relations along with their properties specify the equations induced by the \msp. In addition, each $\aworldmodel$ is associated with a \defn{reference variable} $\refexpr$, which points to the variable in $\aworldmodel$ that holds the correct answer to the question as stated in $\question$. 
We consider each $\aproblem$ to be associated with some structure $(\aworldmodel, \refexpr)$. 

We say that $\aworldmodel$ is \defn{faithful} if it represents the semantics of the problem text according to the framework of $\formalismname$. Further, $\aworldmodel$ is \defn{complete} if $\refexpr$ can be solved with the equations induced by $\aworldmodel$.
A complete world model is \defn{correct} if, when evaluated, $\refexpr$ gives the correct answer to the problem. See \cref{fig:overview} for an example of a world model. \looseness=-1

In order to allow for incremental parsing, we segment the world models into sentence-level logical forms $\lform{i}$, $i = 1, \dots, n$.
The logical form is a sequence that represents the containers and/or relations associated with the corresponding sentence.\footnote{A logical form may be empty. Such will be the case for text outside the coverage of \formalismname.} We can convert $\left(\lform{1},\dots, \lform{n}\right)$ to a world model graph, and vice versa. The two representations are nearly equivalent, with the exception of a few caveats (see \cref{app:conversion} for details). There is no bound on the problem length and, by extension, the number of logical forms. $\formalismname$ is thus
able to represent problems of any arbitrary number of reasoning steps. The assignment of logical forms may be ambiguous in the sense that there may be multiple faithful logical forms for a given sentence (discussed in \cref{sec:ambiguity}).

We consider subgraphs $\asubwm{i}$, for sentence $i$, of the final graph $\aworldmodel$. 
A subgraph $\asubwm{i}$ corresponds to the logical forms up to sentence $i$, i.e., $(\lform{1},\dots,\lform{i}) \mapsto \asubwm{i}$. 
We refer to the subgraph for some sentence index $i$ as the \defn{state} of $i$.
As an example of how world models are built incrementally with states, consider \cref{fig:overview}. The first sentence maps to the container for label \emph{Will} holding the entity \emph{money} of quantity $83$ with unit \emph{dollar}. The second sentence provides information on an update to Will's possessed money, a \transfer relation (\cref{sec:transfer}). Finally, the question sentence introduces rate information, a \rate relation (\cref{sec:rate}), between money and toys.

\saveforcameraready{In the next sections, we describe the details of containers and relations in depth. }

\subsection{Containers}
\label{sec:containers}
We adopt and modify the containers described in the model of \citet{hosseini-etal-2014-learning}. Semantically, containers represent containment/possession. We refer to the possessor in the text as the \defn{label} of the container.\footnote{There may not always be an explicit possessor expressed in text. In such cases, we use the label \emph{World}.} In \cref{fig:overview}, the container label is \emph{Will} for all containers (although in general the label can vary across containers). 
The label must be a noun plus any associated noun adjuncts (like \emph{elementary school}).
In addition to label, a container may have the following four properties:

\paragraph{Entity:} The entity is \emph{what} is possessed in the container. It is a noun, for which there may be an associated count. When expressed in a problem text, it must be the head of a noun phrase. In \cref{fig:overview}, \emph{money} and \emph{toy} are entities.\saveforcameraready{\footnote{Note how the term \emph{money} is not actually expressed in the problem text. Similarly, the word \emph{time} will seldom be expressed in \msps involving reasoning about time. }}
\paragraph{Quantity:} The quantity is the number associated with the entity. It may be known, in which case it will be a positive real number, or unknown, in which case it will be a variable. 
\paragraph{Attribute:} The attribute is a modifier for the entity. It is often an adjective, but may take other forms as well. 
The attribute is an optional property.
\paragraph{Unit:} The unit is the unit of measurement for the entity. A unit property must exist if the entity is a mass noun, but may exist in other cases as well. For example, ``liter of water'' and ``kg of apples'' will both be assigned to containers with units. The unit is an optional property. 

\saveforcameraready{Entity, attribute and unit are written in their lemmatized forms. The label is not, in order to be able to distinguish between a set (plural: \emph{friends}) and an element of some set (singular: \emph{friend}).}

Note that the containers take a variable number of properties; having arity $3$, $4$ or $5$. Two containers are \defn{equal} if they have the same arity and the same properties. We refer to a container's \defn{structure} as its container label, entity, attribute (if exists) and unit (if exists). Two containers are \defn{structurally equal} if they have the same structure. 

\subsection{Relations}
\label{sec:relations}

Relations are the edges in $\aworldmodel$. They represent the interactions between the various parts of the world model, from which the equations of the \msp are induced. The relations are directed, and the direction encodes semantics of the relation depending on the type of relation. Like containers, relations have properties. The properties and their arity also depend on the type of relation. \looseness=-1

There are four types of relations: \transfer, \rate, \explicit and \partwhole. 
Together they span all four basic arithmetic operators $(+,-,\times,\div)$.
Next, we give a detailed description of each of these relation types. Examples of world models with each relation type are provided in \cref{sec:app-examples}.

\subsubsection{\transfer}\label{sec:transfer}
\transfer relations model that a transfer of some quantity of an entity has occurred. A given container structure will either gain or lose quantity from a \transfer relation. For example, ``Alice ate 3 apples'' will correspond to a \transfer with a loss of 3 apples for the container labeled Alice. A \transfer is always between two containers of the same structure. The direction of the edge describes order: The source container will hold the quantity \emph{before} the transfer event occurred, and the target container will hold the quantity \emph{after} the transfer event occurred.

In addition to quantity, \transfer takes the following two properties:
\paragraph{Recipient:} The label of the container structure where the quantity of the given entity is \emph{gained}.
\paragraph{Sender:} The label of the container structure where the quantity of the given entity is \emph{lost}. 

A recipient, a sender or both must exist. \saveforcameraready{\textsc{Transfer} thus has arity $2$ or $3$.} The \transfer relation either adds or subtracts the relation quantity to/from the source container quantity, depending on whether the relation connects the recipient containers or sender containers.

\subsubsection{\rate}\label{sec:rate}
The \rate relation models mathematical rate between two quantities. These two quantities are held in two separate containers with the same label, and the ratio quantity of the rate is given as a property to the relation. \textsc{Rate} has this one single property. 
The direction of the edge determines the relationship: The source container holds the numerator of the rate, and the target container holds the denominator of the rate. In the example in \cref{fig:overview}, the source container holds the entity \emph{money} and the target container holds the entity \emph{toy}, indicating that the rate quantity concerns \emph{money per toy}. Mathematically, \rate implies that the source quantity divided by the relation quantity equals the target quantity.

\subsubsection{\explicit}
\comparison is invoked when there is an explicit relationship between two quantities in the \msp. For example, ``Alice is twice as old as Bob''. \saveforcameraready{The \explicit relation may be either between containers with different labels, such as ``Alice has 3 more apples than Bob'', or between containers with the same label, such as ``Alice has 3 more red apples than she has green apples''.}
It takes two properties; quantity and type:
\paragraph{Type:} The arithmetic operation type \explicit. It can take one of the two values; \emph{add} (indicating addition) or \emph{mul} (indicating multiplication). 

The quantity held in the source container is the one that is combined with the quantity of the \comparison relation under the arithmetic operator, the output of which will be the quantity held in the target container.

\subsubsection{\partwhole}\label{sec:part-whole}
\partwhole relations model set partitions. The set represented by some container is partitioned into subsets, each of which is represented by another container.
For each of the subset containers (the parts), there is an outgoing edge to the container with the superset (the whole). Thus, \partwhole implies that for a given container that has ingoing \partwhole edges, the sum over the quantities in the source containers of those edges equals the quantity in the target container.
Note that \partwhole differs from the other relations in that it requires multiple edges to induce an equation.\footnote{Note that a \partwhole relation can be equivalently represented as a hyperedge.
} In most cases, all containers involved in a \partwhole relation will have the same label. The relation can then be viewed as a relation between entities possessed by a specific label. For instance, ``Alice has 3 red apples and 6 green apples, how many apples does she have in total?'' would be represented by \partwhole. \partwhole relations have no properties.

\partwhole relations may represent meaning that is not explicit in text. Parsing the text of a problem that requires \partwhole might thus lead to an incomplete (\cref{sec:world-model})
world model, which may require additional assumptions.
In addition, orienting \partwhole relations might require commonsense knowledge. For instance, a problem might introduce a quantity for tables and a quantity for chairs, and ask about the total number of furniture. 

\subsection{World model equivalence and similarity} \label{sec:equivalence-similarity}
One of the principal utilities of \formalismname is to allow for evaluating models on their reasoning ability. For that we need consistent equivalence notions and similarity metrics between world models, which we provide here.

Let $\aworldmodel$ and $\aworldmodel'$ be \defn{isomorphic} if there exists an isomorphism on the underlying graphs that additionally preserves relation types. 
We consider two forms of equivalence notions between world models, which we call strong and weak equivalence. Weak equivalence deems two world models to be equal if they are isomorphic. Strong equivalence additionally requires all properties of the containers and relations to be equal.\footnote{In practice, we lemmatize all properties before performing this equivalence check.} In addition, we create two similarity scores based on the AMR metric \emph{smatch} \citep{cai-knight-2013-smatch}: Weak smatch considers graph topology in the same way as our isomorphism equivalence, and strong smatch additionally considers all properties of the world models. We give details on these similarity scores in \cref{sec:app-smatch}.

\subsection{Comparison to other logical formalisms}
\label{sec:comparison_formalisms}

$\formalismname$ can be fully expressed in first-order logic (FOL). We provide a constructive proof in the form of a conversion in \cref{sec:app-fol}, which enables comparison of the expressive power of $\formalismname$ with that of other formalisms.
Both AMR and $\formalismname$ restrict the expressivity of full FOL in different ways. AMR provides a way to express negation (the \texttt{polarity} relation) but does not provide a way to directly express universal quantification\saveforcameraready{\footnote{It is possible to do so indirectly, as in $\neg\exists x.\neg\phi(x) \equiv \forall x. \phi(x)$, but this can only be done once per sentence.}} \citep{bos-2016-squib}. $\formalismname$ represents sets of objects as containers and enables universal quantification over those sets. This is restricted, however, as $\formalismname$ does not allow the definition of sets of sets, or nested universal quantification.\footnote{This disallows higher-order expressions, e.g., \comparison relations between quantities expressed in \transfer relations. It also disallows nested possession outside of what is made possible under \rate, e.g., structures like ``Alice has a house that has a shelf that has a book that has 200 pages.'' } Negation is not directly expressible in $\formalismname$, as it is designed for the domain of \msps where negation is quite rare. \looseness=-1

$\formalismname$ is more comparable to \emph{situation calculus} \citep{McCarthy1963SituationsAA}, where each relation can be modeled as an action that changes the state of the world. Like situation calculus, the changing world state over time is implicitly represented in $\formalismname$ (via the \textsc{Transfer} relation), whereas in FOL, an explicit description of the time of each event is necessary. 

\section{Data Collection}\label{sec:data}

In order to study how models are able to answer \msps,
convert them to logical form, perform world modeling, and reason mathematically to find the answer, we require a diverse dataset of labeled \msps that spans all concepts covered by $\formalismname$.
To ensure diversity and wide variety in the examples, we collect them from numerous sources:
\begin{enumerate}[topsep=2.5pt,noitemsep,leftmargin=13.4pt]
    \item The math word repository \textsc{MAWPS} \citep{koncel-kedziorski-etal-2016-mawps} gathers several datasets \citep{hosseini-etal-2014-learning, kushman-etal-2014-learning, koncel-kedziorski-etal-2015-parsing, roy-roth-2015-solving}, thus providing a wide variety of \msps.
    \item To complement with more challenging problems, we also adopt problems from \textsc{ASDiv-A} \citep{miao-etal-2020-diverse}, which was designed for linguistic diversity and math concept diversity.
    \item We also annotate a subset of the \textsc{SVAMP} dataset \citep{patel-etal-2021-nlp}, which was introduced as a challenge set to test robustness to data artifacts. This enables future work to test the robustness of \formalismname parsers.
\end{enumerate}
We randomly sample a subset from each of these three datasets,\footnote{We also considered the larger \textsc{GSM8K} dataset \citep{cobbe_21_gsm8k}, which contains problems with more reasoning steps. However, although we found $\formalismname$ to cover many of its \msps, annotation workers were unable to reliably annotate these problems. Future work may aim to augment the data to assign ground truth world model structures to longer \msps, using techniques similar to those demonstrated in \cref{sec:generation}.
} and annotate them with world models. We obtain $1,019$ \msps, which corresponds to $3,204$ logical forms, which we partition into $80/20$ train/test splits. \cref{table:data-summary} provides more details. \looseness=-1

\begin{table}[t]
    \fontsize{11}{11}\selectfont
    \centering
    \renewcommand{\arraystretch}{1.75} 
    \setlength{\tabcolsep}{0.5em} 
\begin{tabular}{l|cc|cc}
                 & \multicolumn{2}{c|}{\textbf{Train}} & \multicolumn{2}{c}{\textbf{Test}} \\
                 & MSPs             & LFs              & MSPs            & LFs             \\ \hline
\textsc{ASDIV-A} & \num{328}        & \num{1052}       & \num{83}        & \num{272}       \\
\textsc{MAWPS}   & \num{312}        & \num{936}        & \num{79}        & \num{235}       \\
\textsc{SVAMP}   & \num{173}        & \num{563}        & \num{44}        & \num{146}       \\ \hline
\textsc{TOTAL}   & \num{813}        & \num{2551}       & \num{206}       & \num{653}      
\end{tabular}
    \caption{Size of annotated dataset in terms of number of MSPs and number of sentence-aligned logical forms (LFs), stratified by dataset of origin and split.
    }
    \label{table:data-summary}
\end{table}

We hire external workers for annotation. Annotation follows three phases: A first training phase where annotators are given several small sets at a time with follow-up discussion sessions, an agreement phase in which all annotators are given the same problems and a final scale-up phase. We use an annotation tool created specifically for this work (shown in \cref{sec:app-tool}). The problems are annotated incrementally sentence-by-sentence, in order to match logical forms to sentences as described in \cref{sec:world-model}. Questions are hidden from annotators until all preceding sentences are completed, in order to avoid bias stemming from having read the question---$\formalismname$ is meant to capture the world model of the problem irrespective of what is asked in the question. Within sentences, we ask annotators to add containers and relations according to the order in which they occur in text. This allows us to write the logical forms according to within-sentence order when creating training data for semantic parsing. We maintain this order with integer IDs that are incremented automatically in the annotation tool.

We performed an agreement analysis of $125$ overlapping \msps, revealing a high agreement rate considering the complexity of the annotation task. Concretely, $61$ out of these $125$ were strongly equivalent (\cref{sec:equivalence-similarity}) across annotators, and $107$ were weakly equivalent (\cref{sec:equivalence-similarity}). Many of the only weakly equivalent annotations were due to ambiguity in the properties (\cref{sec:property-ambiguity}), and almost half of the $18$ non-agreed problems were due to ambiguity in relation type (\cref{sec:structural-ambiguity}). The strong and weak smatch scores were $0.91$ and $0.97$ respectively. These can be interpreted as approximate upper bounds on the smatch scores achievable by any model,
due to the ambiguity in the dataset. Many of the annotation errors, also outside of the overlapping set, could be either corrected or discarded \emph{ex post}. Further details on 
annotation are given in \cref{sec:app-annotation}.

\section{Applications of \formalismname}
\label{sec:applications}

In this section we showcase some applications of \formalismname: solving (\cref{sec:solver_framework}), probing of reasoning (\cref{sec:probing}) and generation of new \msps (\cref{sec:generation}).

\subsection{Parsing and Reasoning}\label{sec:solver_framework}
We spell out a framework for solving \msps using \formalismname. 
The framework consists of two components: A \emph{parser} and a \emph{reasoner}. The parser is tasked with assigning a faithful world model $\aworldmodel$ to an input problem $\aproblem$, along with a reference variable $\refexpr$. The reasoner is then queried  
with $\refexpr$ and computes an answer based on the induced equations of $\aworldmodel$. 
We also present a set of initial experiments, meant to introduce the task of \formalismname parsing to the community. 

\subsubsection{Parser}
Given an \msp $\aproblem$, the task is to assign a world model $\aworldmodel$. The first step is to predict the sequence of logical forms $\lform{1},\dots,\lform{n}$. We model this as a conditional distribution 
\begin{align}
    p(\lform{1},\dots,\lform{n} \mid \aproblem) 
    & = \prod_{i=1}^{n} p(\lform{i} \mid \problemelem{1}, \dots, \problemelem{i}).
\end{align}
With this factorization, we can parse the graph incrementally one sentence at a time. The factorization is based on two assumptions: $\lform{i} \perp \problemelem{j}, \forall i < j$ and $\lform{i} \perp \lform{j}, \forall i \neq j$. Both are aligned with $\formalismname$ as outlined in \cref{sec:world-model}: the first assumption means that a logical form is independent of the sentences in subsequent steps, and the second assumption means that logical forms are independent of each other.
Dependencies of logical forms on preceding sentences are kept due to coreferences, elliptical constructions and other inter-sentence dependencies. \looseness=-1

As explained in \cref{sec:world-model}, the logical forms are linearized representations of the world model graphs. Thus, our pipeline (as well as applications like those demonstrated in \cref{sec:applications}) requires that we are able to convert from one representation to the other: World model graphs must be converted to logical forms in order to create training data for a semantic parser, and the predicted logical forms must be converted to world model graphs and reference variables for visualization and reasoning. The details of this conversion are given in \cref{app:conversion}.

\subsubsection{Reasoner}

Once we have a world model graph, we apply a reasoning algorithm over the graph to compute an answer. 
The reasoner takes a world model and a reference variable, and outputs a numeric value for the reference variable $\refexpr$. Our implementation is deterministic and follows two steps. First, it extracts all equations induced by the world model (as described in \cref{sec:relations} and illustrated in \cref{sec:app-examples}). Second, it solves for $\refexpr$ using a recursive algorithm. 
Full pseudocode along with a discussion is presented in \cref{sec:app-reasoner}.\saveforcameraready{\footnote{We note that annotated world models are not necessarily complete (def. in \cref{sec:world-model}). Annotators were requested to only build world models that represent what is made explicit in the text. Some problems may require additional background knowledge to build a complete world model. }}\looseness=-1

\subsubsection{Baseline solving experiments
}\label{sec:parsing-experiments}
We demonstrate our proposed modeling framework with a baseline semantic parser, in the form of a large language model that is supervised in-context. We use Codex \citep{codex}, as language models trained on code have been previously shown to perform well on structured prediction tasks \citep{madaan-commonsense-22, drozdov-compositional-22}. 
The prompt contains 50 ground truth examples from \mawps and \asdiv, and we evaluate the model on the test sets of \mawps, \asdiv and \svamp. We also implement a rule-based baseline system, based on \citet{hosseini-etal-2014-learning}.

Our results corroborate that this is a challenging task; for the least difficult dataset the model gets roughly one third of the problems correct, and predicts a complete world model for only slightly more than half of the problems. The rule-based baseline gets nearly no problems correct. Indeed, a model must, for each sentence, produce well-formed logical forms that exhaustively and correctly capture the semantics in \formalismname, combine these into a world model and query the reasoner with the correct reference variable. One mistake in any of these steps may lead to an incorrect answer.
With much progress and research interest in semantic parsing in recent years \citep{shin-etal-2021-constrained, qiu-etal-2022-improving}
there are several promising directions for improvement.
Further details on the setup and results can be found in \cref{sec:msp_solving_details}.

\begin{figure}[t!]
     \centering
     \includegraphics[width=1.00\linewidth]{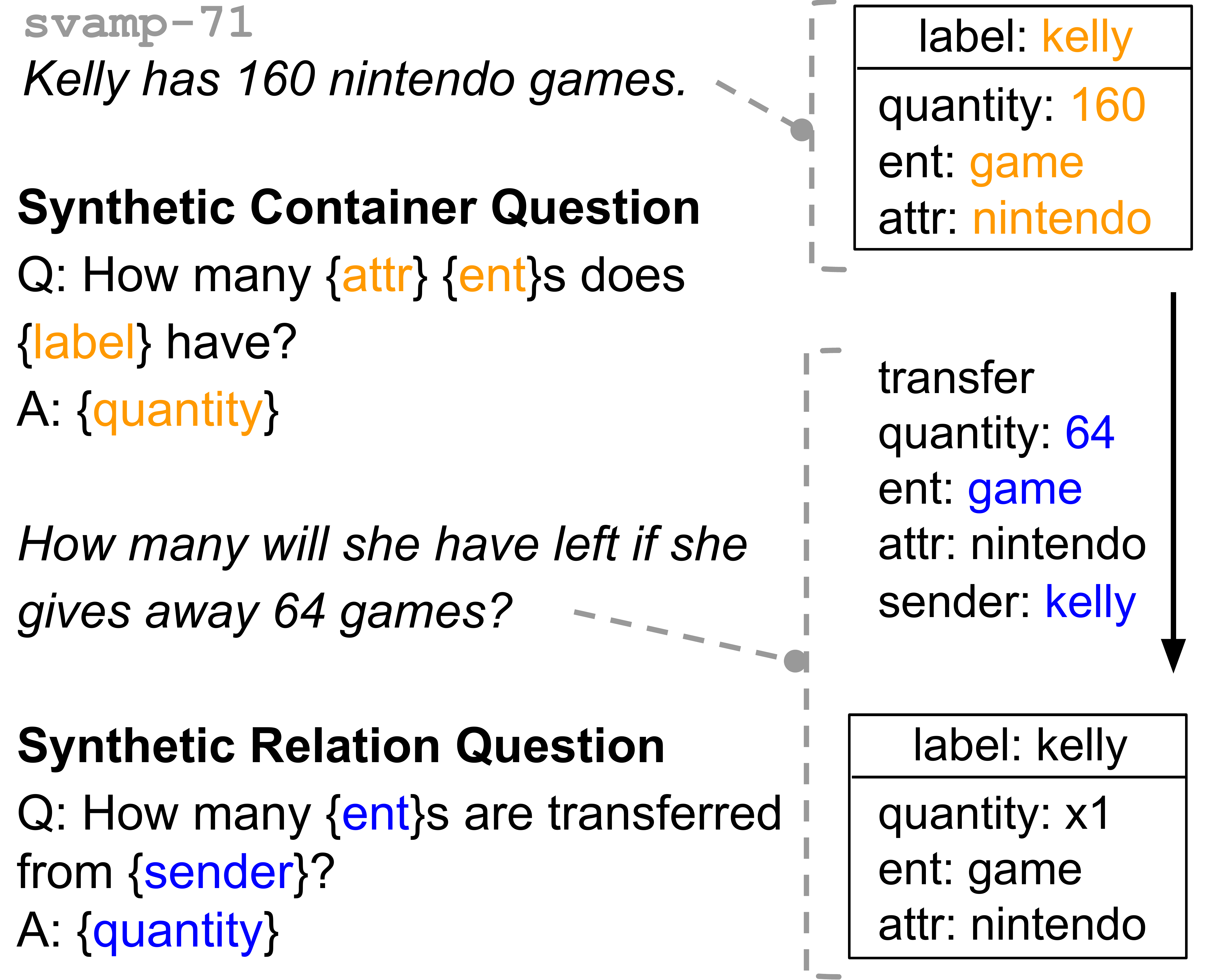} 
     \caption{Synthetically created question-answer pairs based on templates. \saveforcameraready{Note that the quantity in the container or relation does not need to be expressed in text, but could be a variable. Such cases test the model’s ability to reason over intermediate quantities.}
     }
     \label{fig:synthetic_prompting}
\end{figure}

\subsection{Probing \llms' partial knowledge}\label{sec:probing}

World models enable us to study the reasoning ability of \llms: Beyond just testing whether a model outputs the correct solution to an \msp, we can test whether the model follows a correct reasoning path and accurately builds world model representations. 

\paragraph{Setup.}
We design question and answer templates that are automatically filled based on information in the world model.
Two examples of such templates are given in \cref{fig:synthetic_prompting} and a list of all templates is given in \cref{sec:synthetic_prompt_templates}. By courtesy of the world model we know the true answer to each of these synthetic questions, enabling us to create prompts with question-answer pairs. 

We experiment with three types of prompts, all displayed with full-length examples in \cref{table:prompt_types}: \textit{(1) synth QA (all at once)}. We first include the complete problem text, followed by synthetic question and answer pairs related to some part of the text. We randomly sample two such pairs; \textit{(2) synth QA (sentence-by-sentence)}. We again sample two question-answer pairs at random, but in this setting they are imputed right after the sentence in which the answer to the question is given;
\textit{(3) original \msp QA}. Under this setting we do not include any synthetic question-answer pairs, only the original text.
All prompts end with the \msp question that we aim to solve followed by ``A:''. 
We study both whether the synthetic questions help the model answer the \msp correctly, and how well the model answers the synthetic questions themselves.

\begin{table}[]
\fontsize{11}{11}\selectfont
\centering
\renewcommand{\arraystretch}{1.35} 
\setlength{\tabcolsep}{0.4em} 
\begin{tabular}{l|cc}
          & \multicolumn{2}{l}{from $x$ \msps} \\
QA type                     & 0           & 1                    \\ \hline
(1) synth QAs (all at once) & 70.8        & 71.8                 \\
(2) synth QAs (sent by sent) & 71.3        & \textbf{78.6}        \\
(3) original \msp QAs       & 69.4        & 70.8                
\end{tabular}
\caption{Results obtained by GPT-3 in answering math story problems reported in accuracy percent. A larger increase in performance is observed when the synthetic question-answer pairs are presented at the relevant part of the text, rather than at the end. \looseness=-1}
\label{table:prompting}
\end{table}

\paragraph{Results.} We report results obtained by GPT-3 \citep{brown-gpt-3} on the combined test set of all three datasets in \cref{table:prompting}. The number of in-context examples is either 0 or 1.
We observe increased performance when including synthetic question-answer pairs, particularly in setting (2) where the questions are imputed at the relevant part of the \msp text. 
We hypothesize that doing so helps guide the reasoning trace of the model, in a similar vein as chain-of-thought prompting \citep{wei_chain_of_thought}. 
Further, we find that GPT-2 \cite{radford_language_2019}, BART \cite{lewis_bart_2020}, Codex \cite{codex}, T5 \cite{raffel_exploring_2020} and NT5 \cite{yang_nt5_2021} overall perform poorly, but benefit from an increase in performance when synthetic question-answer pairs are provided.

We further compare the ability of GPT-3 to answer the intermediate synthetic questions to its ability to answer the original final question. For each \msp, we first select a container or relation uniformly at random and then create a synthetic question. We then ask both the synthetic question and the original question at the end of two separate prompts in a zero-shot setting.
\cref{table:original_synthetic} displays the results.
Interestingly, in more than one third of the cases that the model gets the original question right (top row), it gets the intermediate synthetic question wrong (top right cell). Overall it also shows a higher accuracy on the original questions (top row) than the synthetic intermediate questions (left column).
While some of these results could be explained by the nature of the templated questions, it does seem to indicate that the model makes use of heuristics rather than human-like reasoning when solving \msps \citep{patel-etal-2021-nlp}.

\begin{table}[]
\fontsize{11}{11}\selectfont
\centering
\renewcommand{\arraystretch}{1.35} 
\setlength{\tabcolsep}{0.4em} 
\begin{tabular}{l|cc}
                  & \multicolumn{2}{c}{Synthetic Question} \\
Original Question & \multicolumn{1}{c|}{Correct}  & Wrong  \\ \hline
Correct           & \multicolumn{1}{c|}{46.0\%}   & 25.7\% \\ \hline
Wrong             & \multicolumn{1}{c|}{11.0\%}   & 17.3\%
\end{tabular}
\caption{We test whether the model gets synthetic questions about parts of the world model right and compare it against its performance on the original question. }
\label{table:original_synthetic}
\end{table}

\subsection{
Generation of \msps}\label{sec:generation}

\formalismname can be considered as a space under which a practitioner can design new \msps with certain desired features. For instance, a teacher may be interested in generating variations of an \msp to test a specific mathematical concept with a specific unknown variable. To demonstrate the potential for such applications we provide a small proof-of-concept experiment. 

\paragraph{Setup.} 
We use GPT 3.5 Turbo \citep{ouyang2022training} with a prompt of 30 examples from the train sets of \mawps and \asdiv. 
One example consists of the logical forms for a full \msp world model (source) followed by the text of the \msp (target). We separate sentence-aligned logical forms in the source as well as the sentences in the target by a marker, so that the model can pick up the alignment patterns. The ground truth examples are sampled randomly. To generate a new \msp conditioned on a world model, we append the logical form corresponding to the world model to the end of the prompt. We try generating new \msps both based on (i) world models present in our annotated test sets (paraphrasing) and (ii) manual augmentations of annotated world models. We perform evaluation for setting (i) using SacreBLEU \citep{post-2018-call} and BERTScore \citep{bert-score}, comparing all \msps in the test sets to their paraphrases.\footnote{More details on the generation setup are given in \cref{sec:generation_details}.}\looseness=-1

\begin{figure}[t!]
     \centering
     \includegraphics[width=1.00\linewidth]{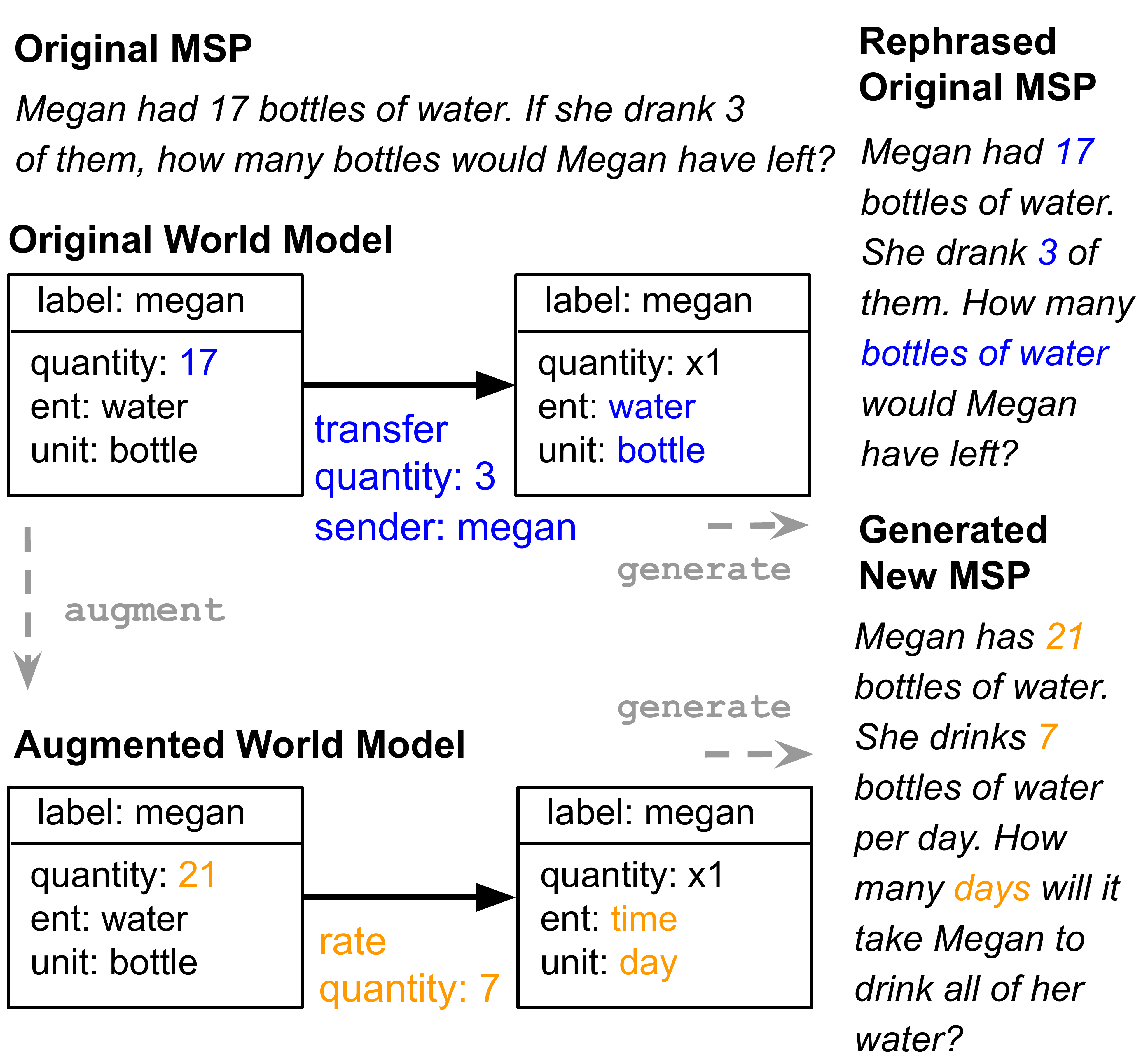} 
     \caption{Example \msps generated by GPT 3.5 Turbo. 
     }
     \label{fig:conditioned_generation}
\end{figure}

\paragraph{Results.} We obtain SacreBLEU scores of $66.73$, $40.86$ and $26.02$ and F1 BERTScores of $0.933$, $0.930$ and $0.931$ for \mawps, \asdiv and \svamp respectively. Qualitatively we observe that the generated \msps mostly stay faithful to the logical forms but tend to be shorter and less linguistically complex than the original problems, which would explain the comparatively low SacreBLEU scores in comparison to the BERTScores.
Further, we give the first six examples we generated according to the described setup. One of them is shown in \cref{fig:conditioned_generation}. The model generates an output \msp very similar to the original, having only accessed the original's ground truth logical forms. We further augment the original world model by changing the \transfer to a \rate. Note how the generated \msp is faithful to the augmented world model. The other five examples are shown in \cref{table:constrained_generation}.

\section{Conclusion}
In this work, we have presented a novel formalism, \formalismname, for expressing the semantics of math story problems. 
We have annotated a \formalismname corpus consisting of $1,019$ problems and $3,204$ logical forms.
A world model derived from \formalismname exposes the structure of the reasoning process needed to solve the problem,
which benefits several applications as we have demonstrated in \cref{sec:applications}.
As such, we hope that \formalismname will promote use cases beyond just improved \msp solving, 
ranging from automated chain-of-thought prompting to math problem generation.

\section*{Limitations}
\formalismname is limited to cover math story problems using the four basic arithmetic operators. Furthermore, within the space of such problems, it does not cover ``second-order'' \msps (as discussed in \cref{sec:comparison_formalisms}). Neither does it cover negation nor inequalities.

We only consider datasets with \msps written in English in this work. However, \formalismname should in principle be able to cover the same type of problems formulated in other languages as well.

An obvious limitation of this work is the low performance on the task of solving MSPs. The focus of this work is to introduce the world model formalism and its use cases, and we leave for future work to build stronger \formalismname parsers.

\section*{Ethics Statement}

We foresee no major ethical concerns with this work. The introduction of \formalismname is aimed at improving the interpretability and robustness of existing and future models for math story problem solving. On this account, we hope to contribute to identifying (and hopefully reducing) existing biases in pre-trained language models, or any future alternatives. However, we would like to caution that the formalism could be used to generate inappropriate math story problems. 

\section*{Acknowledgements}

We thank Arnav Mishra, Aryaman Kolhe, Devraj Thakur, Gaurav Saini and Soham Bopardikar for help with annotation work.  
We further thank Jakub Macina, Kumar Shridhar and Menna El-Assady for input in the early stages of the project, Ethan Wilcox and Ying Jiao for helpful feedback, and Yixiong Wang for help in implementation of a symbolic baseline solver.
Andreas Opedal is partially supported by the Max Planck ETH Center for Learning Systems.
Niklas Stoehr acknowledges funding from the Swiss Data Science Center.

\bibliography{references/custom}
\bibliographystyle{acl_natbib}

\appendix

\section{$\formalismname$ Examples}
\label{sec:app-examples}

\subsection{\transfer}

Consider the following problem:

\vspace{0.6em}
\begin{minipage}{19em}
  The school cafeteria had 14 apples. If they used 13 to make lunch for the students and then bought 49 more, how many apples would they have?
\end{minipage}
\vspace{0.3em}

\begin{figure}
    \centering
    \includegraphics[width=0.45\textwidth]{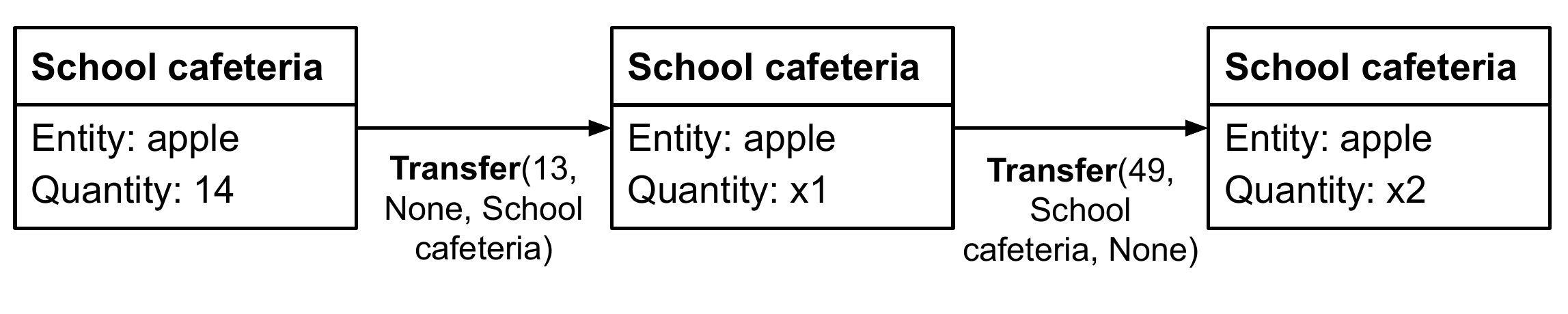}
    \caption{Example of a world model using \textsc{Transfer}.}
    \label{fig:world-model-school-cafeteria}
\end{figure}

We display the corresponding world model in \cref{fig:world-model-school-cafeteria}. The first sentence will correspond to a container for \emph{school cafeteria} that holds $14$ of entity \emph{apple}. The second sentence describes two transfers: a first one where the school cafeteria is the sender of $13$ apples, and a second one where the school cafeteria is the recipient of $49$ apples. We get two equations:
\begin{align}
    14-13=x_1 \\
    x_1+49=x_2
\end{align}
The question asks for how many apples the school cafeteria has in the end, which matches the container holding the variable $x_2$ in the world model.

\begin{figure}
    \centering
    \includegraphics[width=0.45\textwidth]{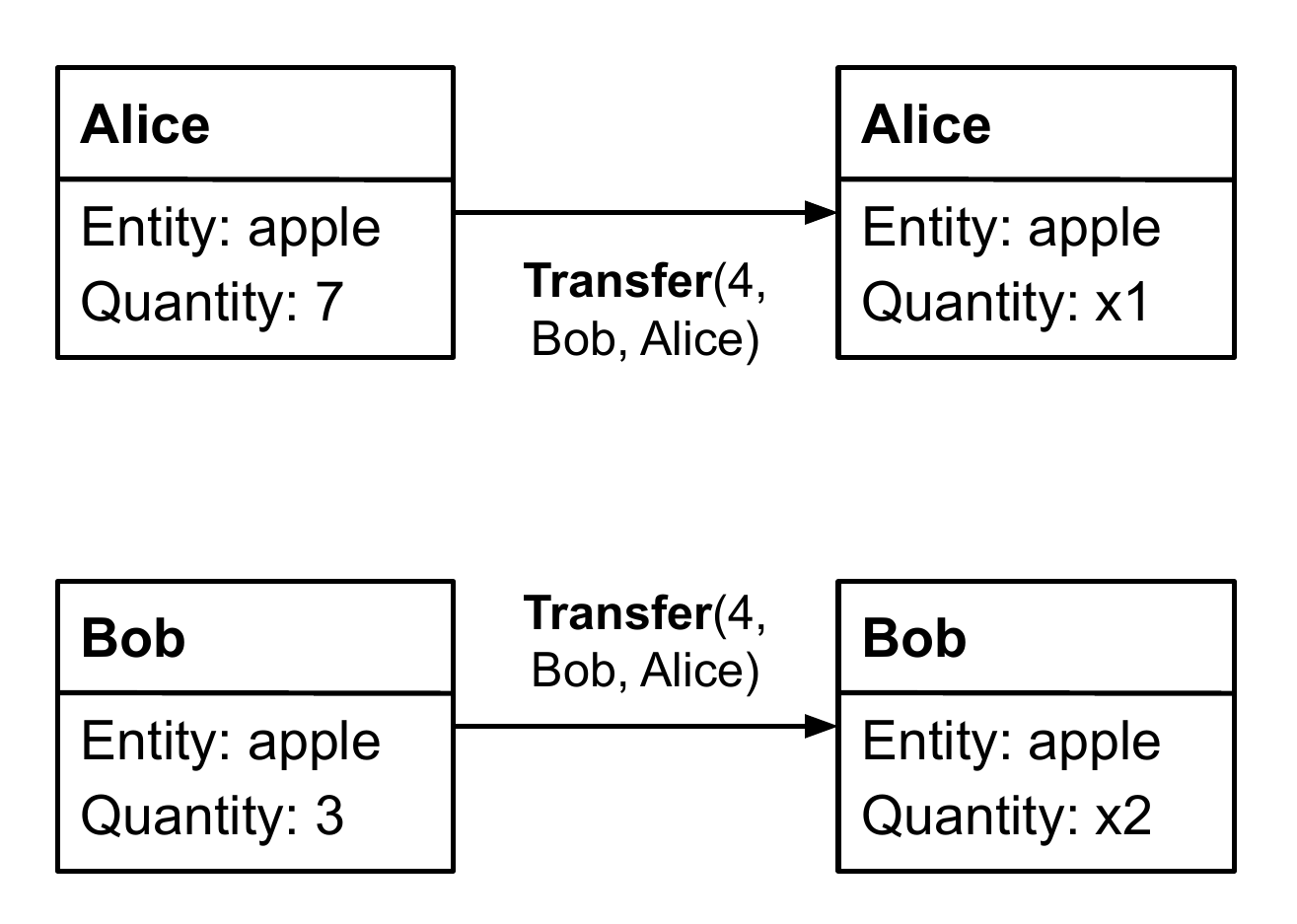}
    \caption{Example of a world model using \textsc{Transfer}.}
    \label{fig:world-model-alice-bob-apples}
\end{figure}

Although the \transfer relation always connects two containers of the same structure in the graph, a transfer event may occur between two containers of different structure. For example, ``Alice gives 3 apples to Bob'' describes a transfer event with Alice losing 3 apples and Bob gaining 3 apples. In these cases, we need two edges with the same properties in the world model; one for Alice's containers and one for Bob's containers (see \cref{fig:world-model-alice-bob-apples}). Consider the following problem with a transfer event occurring between two different possessors:

\vspace{0.6em}
\begin{minipage}{19em}
  Alice has 7 apples and Bob has 4 apples. Alice gives 3 apples to Bob. How many apples does Bob have now?
\end{minipage}
\vspace{0.3em}

We show the corresponding world model in \cref{fig:world-model-alice-bob-apples}. \emph{Alice} and \emph{Bob} are represented by two separate containers, which are both updated by the same transfer event.

\subsection{\rate}
Consider the following problem:

\vspace{0.6em}
\begin{minipage}{19em}
  Lansing has 25 elementary schools. There are 247 students in each school. How many elementary students are there altogether in Lansing?
\end{minipage}
\vspace{0.3em}

This is a rate problem, as we get a rate on the number of students per elementary schools in the second sentence. The relation induces the following equation:
\begin{align}
    \frac{x_1}{25} = 247
\end{align}
The question asks for the total number of students in Lansing, which corresponds to the quantity in the container that holds the entity \emph{student}.

\begin{figure}
    \centering
    \includegraphics[width=0.45\textwidth]{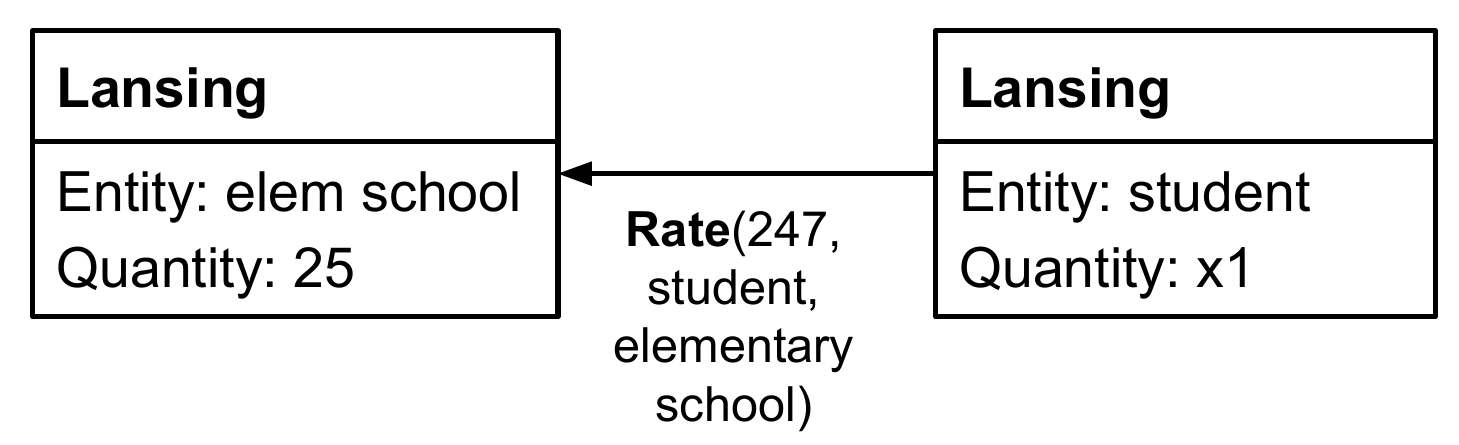}
    \caption{Example of a world model using \textsc{Rate}.}
    \label{fig:world-model-lansing}
\end{figure}

\subsection{\explicit}
Consider the following problem:

\vspace{0.6em}
\begin{minipage}{19em}
  James has 232 balloons. Amy has 101 balloons. How many more balloons does James have than Amy?
\end{minipage}
\vspace{0.3em}

\begin{figure}
    \centering
    \includegraphics[width=0.45\textwidth]{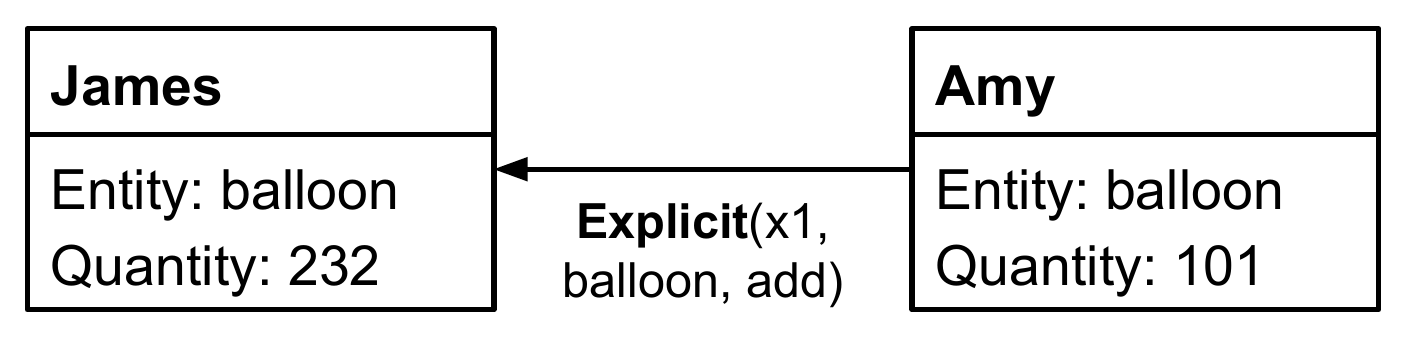}
    \caption{Example of a world model using \comparison.}
    \label{fig:world-model-balloon}
\end{figure}

The first two sentences will correspond to two containers, representing the number of balloons possessed by James and Amy respectively. In the question sentence, we get information about an \comparison relation between these two containers, with properties $x_1$ and \emph{add}. Since we need to add the balloons in Amy's container to get the number of balloons in James' container, the edge is directed outwards from Amy's container. This relation induces the following equation:
\begin{align}
    101+x_1=232
\end{align}
The world model is displayed in \cref{fig:world-model-balloon}.

\subsection{\partwhole}
Consider the following problem:

\vspace{0.6em}
\begin{minipage}{19em}
  Gavin has 23 shirts. 6 are blue the rest are green. How many green shirts does Gavin have?
\end{minipage}
\vspace{0.3em}

\begin{figure}
    \centering
    \includegraphics[width=0.45\textwidth]{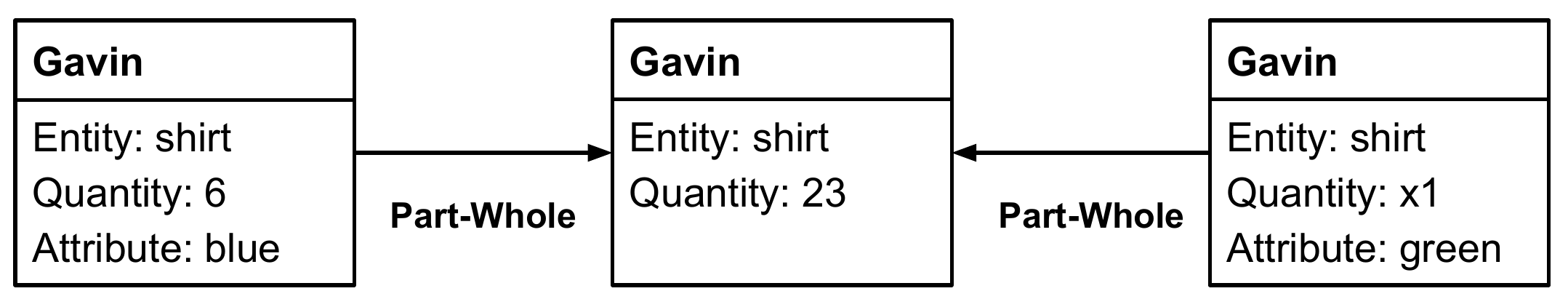}
    \caption{Example of a world model using \textsc{PartWhole}.}
    \label{fig:world-model-shirts}
\end{figure}

The first sentence will correspond to a container for Gavin holding the quantity of his shirts. The part-whole information is introduced in the second sentence, in which the $6$ refers to shirts in the previous sentence (via an elliptical construction), and ``the rest'' tells us we have an additional complementing part of green shirts. Hence, the second sentence is assigned two new containers with attributes \emph{blue} and \emph{green}, as well as \textsc{PartWhole} relations from both of these containers to the whole container introduced in the first sentence. This leads to the following equation:
\begin{align}
    6+x_1=23
\end{align}
The reference variable is the quantity in the container holding Gavin's green shirts. See \cref{fig:world-model-shirts} for the world model.

\section{Ambiguity}
\label{sec:ambiguity}
Ambiguity occurs when the same problem text may be assigned multiple correct and faithful world models. We distinguish between two types of ambiguity for $\formalismname$: \defn{property ambiguity} and \defn{structural ambiguity}.

\subsection{Property ambiguity} \label{sec:property-ambiguity}
Property ambiguity concerns cases where there are multiple possible properties to containers and/or relations that yield a semantically faithful world model. For instance, it is ambiguous whether ``carrot sticks'' is to be interpreted as an entity \emph{carrot stick}, as entity \emph{carrot} with unit \emph{stick}, or as entity \emph{stick} with attribute \emph{carrot}. Property ambiguity may also follow from syntactic ambiguity in the problem text.

\subsection{Structural ambiguity} \label{sec:structural-ambiguity}
Structural ambiguity occurs when the topology, including relation types, differs between several correct and faithful world models for a given problem. Consider the following example:

\vspace{0.6em}
\begin{minipage}{19em}
  James ate 22 carrot sticks before dinner and 15 more after dinner. How many carrot sticks did he eat?
\end{minipage}
\vspace{0.3em}

\begin{figure}[h]
    \centering
	\includegraphics[width=0.45\textwidth]{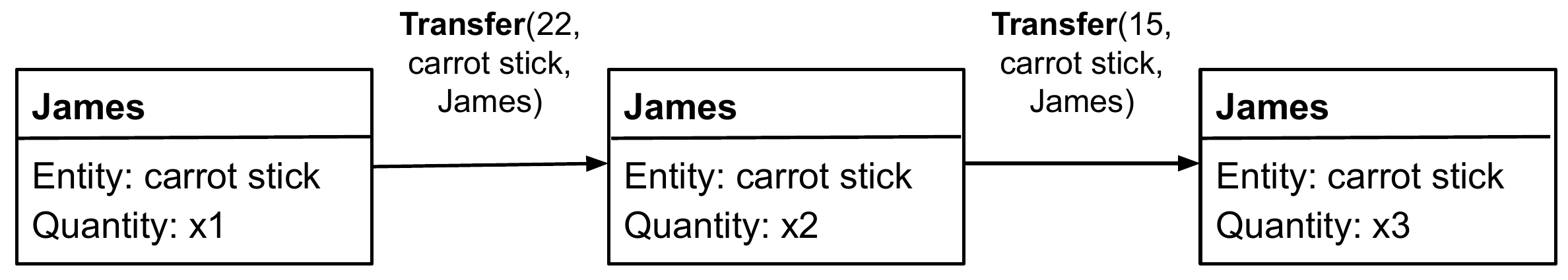}
	\caption{World model with a Transfer interpretation.}
	\label{fig:carrot-stick-transfer}
\end{figure}

This problem could be modeled either with \textsc{Transfer} or \textsc{PartWhole}. In the case of \textsc{Transfer}, we view James as possessing some quantity of carrot sticks to start with. He then eats $22$ of these, which can be viewed as a \textsc{Transfer} where James is the sender. This \textsc{Transfer} relation will be an outgoing edge into a new updated container for James' carrots. Another \textsc{Transfer} occurs for the $15$ carrot sticks he ate after dinner. The reference variable would then be the variable held in the first container -- how many carrot sticks James had initially. See \cref{fig:carrot-stick-transfer} for the world model. Note that such a world model is not sufficient for solving the problem without further assumptions, it requires defeasible reasoning \citep{sep-reasoning-defeasible}. We must assume that James had no carrot sticks after having eaten the ones post dinner, corresponding to the third container holding quantity $0$, in order for the world model to be complete.\footnote{An alternative would be to augment $\refexpr$ to handle expressions, giving $\refexpr=22+15$. This would involve a more complex linarization scheme than that described in \cref{sec:app-linearization} however.}

\begin{figure}[h]
    \centering
	\includegraphics[width=0.45\textwidth]{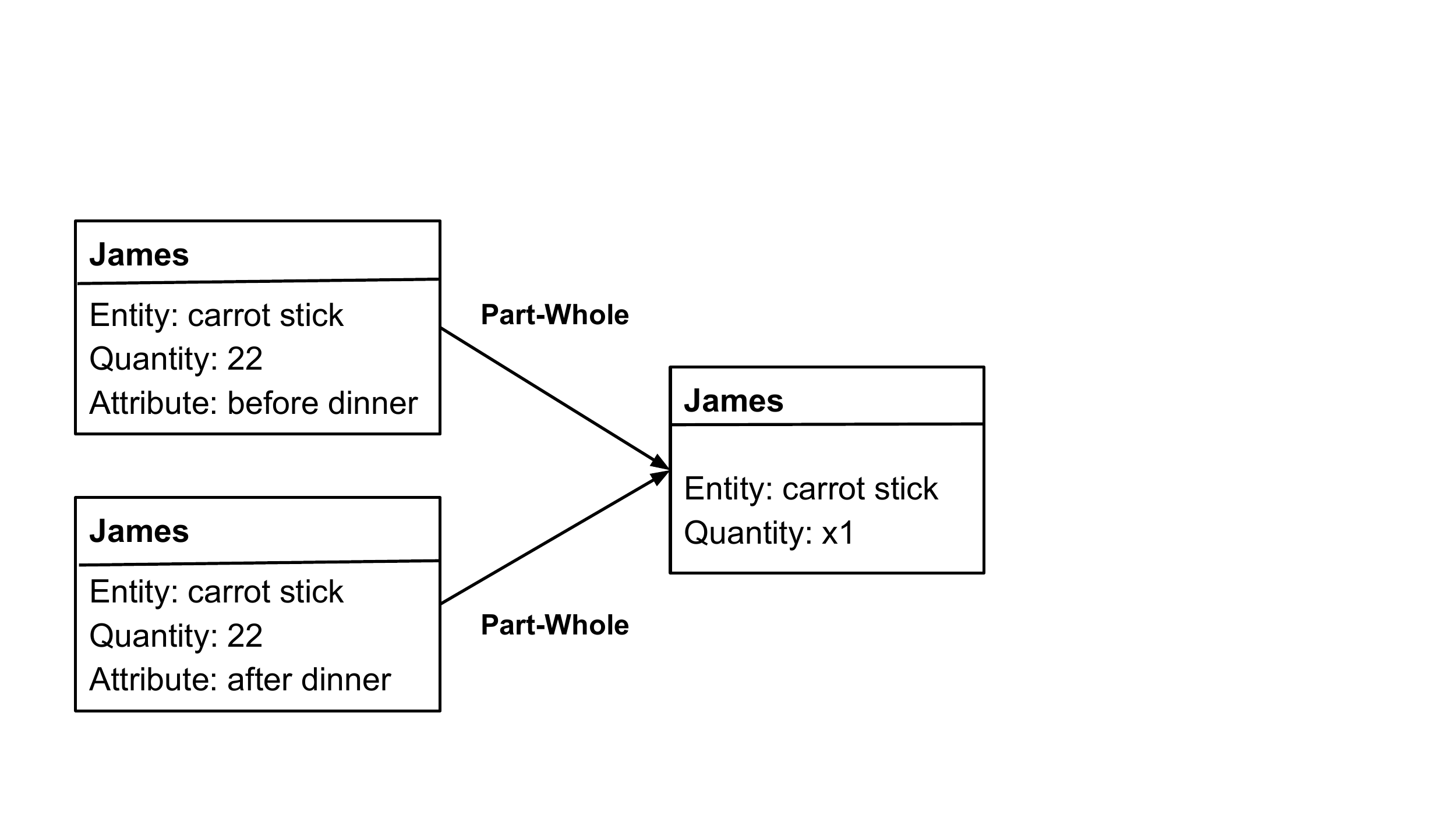}
	\caption{World model with a part-whole interpretation.}
	\label{fig:carrot-stick-part-whole}
\end{figure}

Another possibility would be with \textsc{PartWhole}. With \textsc{PartWhole}, we take the static view of James possessing $22$ carrot sticks before dinner and $15$ carrot sticks after dinner, assigning a container for each. The question statement gives us the information that we are asking for the total number of carrot sticks, which would be parsed with \textsc{PartWhole} to a container with the total. The reference will refer to the variable in this latter container. In contrast to the \textsc{Transfer} interpretation, the \textsc{PartWhole} interpretation does not require additional assumptions to create a complete world model. See \cref{fig:carrot-stick-part-whole}.

\section{Similarity Scores}
\label{sec:app-smatch}
In this section, we describe how we adapt smatch \citep{cai-knight-2013-smatch} for measuring similarity between world model graphs. We express the world models as conjunctions over logical triples. We label all containers and relations with a unique variable, and denote that such a variable is an instance of a container or one of the five relation types with the triple $\texttt{instance(variable, type)}$. Containers are represented as arguments to the relations in the form of $\texttt{source}$ and $\texttt{destination}$, which are non-core roles in AMR.\footnote{We refer to the AMR guidelines for more information: \href{https://github.com/amrisi/amr-guidelines}{https://github.com/amrisi/amr-guidelines}} For instance, a container $\texttt{c}$ being the source node of relation $\texttt{r}$ is represented as $\texttt{source(r, c)}$. The topology smatch score of two world models is then computed by taking the maximum f-score over one-to-one variable mappings between the two world models, as in \citet{cai-knight-2013-smatch}. 

The full semantic smatch score is computed in the same way, with the addition of logical triples for all the container and relation properties. We define core argument roles for the containers and each of the relation types. For instance, $\texttt{ARG0}$ of a container will be its entity. The entity \emph{apple} belonging to container $\texttt{c}$ will be represented by two logical triples $\texttt{instance(e, apple)}$ and $\texttt{ARG0(c, e)}$. 

\section{Conversion to First-order Logic}
\label{sec:app-fol}

In this section, we define a function to convert world model graphs into an equivalent FOL expression.

\subsection{Describing quantities}
Before introducing the conversion function, we first present a way in which quantities are described in FOL, as a preliminary. We define the \texttt{Measure} predicate, which is used to describe the ``size'' of a set. The set may contain countable entities such as ``8 balloons'' or uncountable entities such as ``10 grams of coffee,'' and \texttt{Measure} is used to specify both types of quantities.

We introduce axioms to enable mathematical reasoning over the \texttt{Measure} predicate. If the measure of a set is a cardinal number (as in ``8 balloons''), then it is the cardinality of that set:
\begin{align*}
    \forall x\forall m(&\texttt{Measure}(x,m) \land m\in\{0,1,\hdots\} \\
    &\hspace{1.5em}\leftrightarrow \texttt{Cardinality}(x,m)).
\end{align*}
For example, if a set $x$ contains $8$ elements, we write $\texttt{Measure}(x,8)$. We also define the additivity of measures:
\begin{align*}
    &\forall x\forall y\forall m_x\forall m_y(x\cap y = \varnothing \\
    &\hspace{1.5em}\Land \texttt{Measure}(x,m_x) \Land \texttt{Measure}(y,m_y) \\
    &\hspace{0.5em}\to \texttt{Measure}(x\cup y, m_x + m_y)).
\end{align*}
That is, for any disjoint sets, the measure of their union is equal to the sum of their measures. To describe the size of sets containing uncountable entities (as in ``10 grams of coffee''), we use the \texttt{Quantity} predicate. For example, if a set $x$ contains $10$ grams, we write $\texttt{Measure}(x,\texttt{Quantity}(10,\texttt{Gram}))$. To enable reasoning over such measures, we define the following axiom:
\begin{align*}
    \forall x\forall y\forall u(&\texttt{Quantity}(x,u)+\texttt{Quantity}(y,u) \\
    &\hspace{1.5em}= \texttt{Quantity}(x+y,u)).
\end{align*}
That is, quantities may be summed if they share the same units. Subtraction of quantities is defined similarly. Further axioms can be defined to allow conversions between units, such as:
\begin{align*}
    \forall x(&\texttt{Quantity}(x,\texttt{Milliliter}) \\
    &\hspace{1.5em}= \texttt{Quantity}(x/1000,\texttt{Liter})).
\end{align*}

\subsection{Conversion function}
Let $g = (V,E)$ be world model graph consisting of a set of containers $V$ (i.e. vertices) and relations $E$ (i.e. edges). Let $\bar{E}\subseteq E$ be the subset of relations that do not have type \textsc{PartWhole} (for which the semantics of the edges are not independent and thus need to be treated separately). Recall that the world model may also contain variables, which represent unknown quantities. Let $U$ be the set of these variables. We can define a function $\|g\|$ that converts $g$ into an equivalent FOL expression.
\begin{align*}
    \|g\| &= \exists v_1 \hdots \exists v_{|V|} \exists e_1 \hdots \exists e_{|\bar{E}|} \exists u_1 \hdots \exists u_{|U|} ( \\
        &\|V_1\| \land \hdots \land \|V_{|V|}\| \land \|\bar{E}_1\| \land \hdots \land \|\bar{E}_{|\bar{E}|}\| ).
\end{align*}

\subsubsection{Converting containers}
Recall that each container in the world model $V_i \in V$ is labeled with a set of properties: the label (denote as $\mathcal{L}_i$), entity ($\mathcal{E}_i$), quantity ($\mathcal{Q}_i$), attribute ($\mathcal{A}_i$), and unit ($\mathcal{U}_i$). Note that the unit property is optional depending on whether the entity $\mathcal{E}_i$ is countable or not. 
If the entity is countable, the container is mapped to a definition of a set:
\begin{align*}
    \|V_i\| &= \texttt{Owner}(v_i,\mathcal{L}_i) \\
        &\hspace{0.5em}\Land \texttt{Measure}(v_i,\|\mathcal{Q}_i\|) \\
        &\hspace{0.5em}\Land \forall x\in v_i(\mathcal{E}_i(x) \Land \mathcal{A}_i(x)) \Land \|E^{\texttt{PW},i}\|,
\end{align*}
where $E^{PW,i}\subseteq E$ is the set of edges of type \textsc{PartWhole} whose target vertex is $i$. Otherwise, if the entity is uncountable:
\begin{align*}
    \|V_i\| &= \texttt{Owner}(v_i,\mathcal{L}_i) \land \mathcal{E}_i(v_i) \land \mathcal{A}_i(v_i) \\
        &\hspace{0.5em}\land \texttt{Measure}(v_i,\texttt{Quantity}(\|\mathcal{Q}_i\|,\mathcal{U}_i)) \\
        &\hspace{0.5em}\land \|E^{\texttt{PW},i}\|.
\end{align*}
Note that the attribute and unit properties may be omitted, and if the container $v_i$ is missing a property, the corresponding conjunct is omitted as well (e.g., if the container is missing an attribute property, the conjunct $\mathcal{A}_i(\cdot)$ is omitted). Each quantity $\mathcal{Q}_i$ is mapped as follows:
\begin{equation*}
    \|\mathcal{Q}_i\| =
        \begin{cases}
            \mathcal{Q}_i, &\text{if } \mathcal{Q}_i \in \mathbb{R}, \\
            u_j, &\text{if } \mathcal{Q}_i = x_j \text{ for some } x_j \in U.
        \end{cases}
\end{equation*}
Unlike other relations, the semantics of \textsc{PartWhole} edges are not independent of each other, and so we define them here as a special case:
\begin{equation*}
    \|E^{\texttt{PW},i}\| = \texttt{PartWhole}(\{v_{s_1},v_{s_2},\hdots\},v_i),
\end{equation*}
where $s_j$ is the index of the source vertex of the edge $E^{\texttt{PW},i}_j$, and so $\{v_{s_1},v_{s_2},\hdots\}$ is the set of the source vertices of the \textsc{PartWhole} edges with target vertex $i$. In section \ref{sec:relation_semantics}, we provide axioms that define the semantics of each relation, including \texttt{PartWhole}. 

\subsubsection{Converting relations}
Each relation $\bar{E}_i \in \bar{E}$ is also converted into a conjunction. Let $s_i$ be the index of the source vertex of $\bar{E}_i$, and similarly let $t_i$ be the index of the target vertex.

If the edge $\bar{E}_i$ is labeled as \textsc{Transfer}, it may have the following properties: the sender (denote as $\mathcal{S}_i$), recipient ($\mathcal{R}_i$), entity ($\mathcal{E}_i$), quantity ($\mathcal{Q}_i$), attribute ($\mathcal{A}_i$), and unit ($\mathcal{U}_i$). Similarly to containers, the entities in relations may be countable or uncountable. For brevity, we only show the conversion for the case where the entities are countable, but the conversion of uncountable quantities mirrors that shown for containers above. In this case, the \textsc{Transfer} edge is converted:
\begin{align*}
    &\|\bar{E}_i\| = \texttt{Transfer}(e_i) \\
        &\hspace{1.0em}\Land \texttt{Source}(e_i,v_{s_i}) \Land \texttt{Target}(e_i,v_{t_i}) \\
        &\hspace{1.0em}\Land \texttt{Time}(v_{s_i})+1=\texttt{Time}(v_{t_i}) \\
        &\hspace{1.0em}\Land \texttt{Sender}(e_i,\mathcal{S}_i) \Land \texttt{Recipient}(e_i,\mathcal{R}_i) \\
        &\hspace{1.0em}\Land \exists r(\texttt{Arg}(e_i,r) \Land \texttt{Measure}(r,\|\mathcal{Q}_i\|) \\
        &\hspace{2.0em}\Land \forall x\in r(\mathcal{E}_i(x) \Land \mathcal{A}_i(x))).
\end{align*}

If the edge $\bar{E}_i$ is labeled as \textsc{Rate}, it may have the following properties: the entity ($\mathcal{E}_i$), quantity ($\mathcal{Q}_i$), attribute ($\mathcal{A}_i$), and unit ($\mathcal{U}$). Then, the edge is converted:
\begin{align*}
    &\|\bar{E}_i\| = \texttt{Rate}(e_i) \\
        &\hspace{1.0em}\Land \texttt{Source}(e_i,v_{s_i}) \Land \texttt{Target}(e_i,v_{t_i}) \\
        &\hspace{1.0em}\Land \texttt{Time}(v_{s_i})=\texttt{Time}(v_{t_i}) \\
        &\hspace{1.0em}\Land \exists r(\texttt{Arg}(e_i,r) \Land \forall y\in r(\texttt{Measure}(y,\|\mathcal{Q}_i\|) \\
        &\hspace{2.0em}\Land \forall x\in y(\mathcal{E}_i(x) \Land \mathcal{A}_i(x)))).
\end{align*}

Finally, if the edge $\bar{E}_i$ is labeled as \comparison, it may have the following properties: the type ($\mathcal{T}_i \in \{\texttt{Add}, \texttt{Mul}\}$), quantity ($\mathcal{Q}_i$), and unit ($\mathcal{U}_i$). Then, the edge is converted:
\begin{align*}
    &\|\bar{E}_i\| = \texttt{Comparison}\mathcal{T}_i(e_i) \\
        &\hspace{1.0em}\Land \texttt{Source}(e_i,v_{s_i}) \Land \texttt{Target}(e_i,v_{t_i}) \\
        &\hspace{1.0em}\Land \texttt{Time}(v_{s_i})=\texttt{Time}(v_{t_i}) \\
        &\hspace{1.0em}\Land \exists r(\texttt{Arg}(e_i,r) \Land \texttt{Measure}(r,\|\mathcal{Q}_i\|) \\
        &\hspace{1.5em}\Land \forall x\in r(\mathcal{E}_i(x) \Land \mathcal{A}_i(x))).
\end{align*}

Note that in the above, the sender, recipient, attribute, and unit properties are optional. If the relation is missing any property, the corresponding conjunct is omitted (e.g., if the attribute property is missing, the corresponding term $\mathcal{A}_i(x)$ is omitted).

\begin{figure}
    \small
    \begin{tabular}{p{0.46\textwidth}}
        \hline
        \vspace{-0.2em}
        \textbf{Natural language representation:} \\
        ``James has 232 balloons. Amy has 101 balloons. How many more balloons does James have than Amy?'' \\[0.35em]
        \hline \\[-0.5em]

        \textbf{\formalismname representation:} \\
        \includegraphics[width=0.47\textwidth]{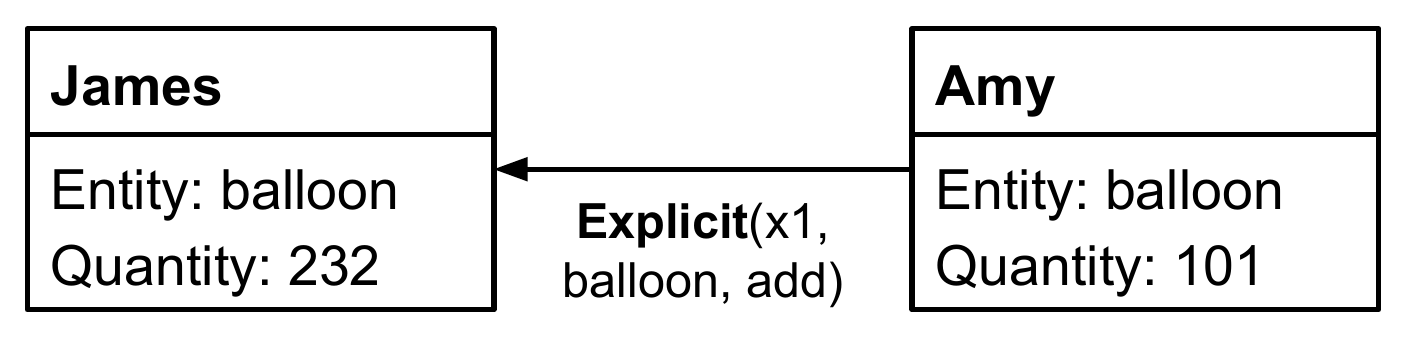} \\[0.2em]
        \hline \\[-0.5em]

        \textbf{First-order logic representation:} \\[-1.8em]
        {\begin{align*}
            &\exists v_1 \exists v_2 \exists e_1 \exists u_1 ( \\
            &\hspace{0.5em}\texttt{Owner}(v_1,\texttt{James}) \\
            &\hspace{0.5em}\Land \texttt{Measure}(v_1,232) \Land \forall x\in v_1.\texttt{balloon}(x) \\
            &\hspace{0.5em}\Land \texttt{Owner}(v_2,\texttt{Amy}) \\
            &\hspace{0.5em}\Land \texttt{Measure}(v_2,101) \Land \forall x\in v_2.\texttt{balloon}(x) \\
            &\hspace{0.5em}\Land \texttt{ComparisonAdd}(e_1) \\
            &\hspace{0.5em}\Land \texttt{Source}(e_1,v_2) \Land \texttt{Target}(e_1,v_1) \\
            &\hspace{0.5em}\Land \texttt{Time}(v_2)=\texttt{Time}(v_1) \\
            &\hspace{0.5em}\Land \exists r( \texttt{Arg}(e_1,r) \Land \texttt{Measure}(r,u_1) \\
            &\hspace{1.5em}\Land \forall x\in r.\texttt{balloon}(x)))
        \end{align*}} \\[-1.8em]
        \hline
    \end{tabular}
    \caption{Example of a math story problem with its equivalent representations as a world model graph and in first-order logic.}
    \label{fig:fol-example}
\end{figure}

See Figure \ref{fig:fol-example} for an example application of the above conversion function.

\subsection{Semantics of relations and predicates} \label{sec:relation_semantics}
We define the semantics of each relation, starting with the \textsc{Rate} relation:
\begin{align*}
    &\forall e\forall v_s \forall v_t \forall r (\texttt{Rate}(e) \Land \texttt{Arg}(e,r) \\
    &\hspace{1.5em}\Land \texttt{Source}(e,v_s) \Land \texttt{Target}(e,v_t) \\
    &\hspace{0.5em}\to \texttt{Partition}(r,v_s) \hspace{0.2em}\Land \\
    &\hspace{1.5em}\exists m(\texttt{Measure}(r,m) \land \texttt{Measure}(v_t,m))),
\end{align*}
where $\texttt{Partition}(r,v_s)$ denotes that $r$ is a \emph{partition} of the set $v_s$: $r$ is a set of disjoint subsets of $v_s$ such that their union is equal to $v_s$. More precisely:
\begin{align*}
    &\forall x\forall y \Big(\texttt{Partition}(x,y) \leftrightarrow \\
    &\hspace{0.5em}\forall z,z'\in x(z\ne z' \to z \cap z' = \varnothing) \Land y = \bigcup_{z\in x} z\Big).
\end{align*}
We also use the notion of a partition to define the semantics of the \textsc{Transfer} relation:
\begin{align*}
    &\forall e\forall v_s \forall v_t \forall r (\texttt{Transfer}(e) \Land \texttt{Arg}(e,r) \\
    &\hspace{1.5em}\Land \texttt{Source}(e,v_s) \Land \texttt{Target}(e,v_t) \\
    &\hspace{0.5em}\to \exists z( \texttt{Owner}(v_s, z) \Land \texttt{Owner}(v_t, z) \\
    &\hspace{2.5em}\Land \texttt{Recipient}(e, z) \\
    &\hspace{2.5em}\Land \texttt{Partition}(\{r,v_s\},v_t)) \\
    &\hspace{1.0em} \lor \exists z( \texttt{Owner}(v_s, z) \Land \texttt{Owner}(v_t, z) \\
    &\hspace{2.5em}\Land \texttt{Sender}(e, z) \\
    &\hspace{2.5em}\Land \texttt{Partition}(\{r,v_t\},v_s))). \\
\end{align*}
We define the semantics of \textsc{ComparisonAdd}:
\begin{align*}
    &\forall e\forall v_s \forall v_t \forall m_s \forall m_t \forall r (\\
    &\hspace{1.5em}\texttt{ComparisonAdd}(e) \Land \texttt{Arg}(e,r) \\
    &\hspace{1.5em}\Land \texttt{Source}(e,v_s) \Land \texttt{Target}(e,v_t) \\
    &\hspace{1.5em}\Land \texttt{Measure}(v_s,m_s) \Land \texttt{Measure}(v_t,m_t) \\
    &\hspace{0.5em}\to m_s+r=m_t).
\end{align*}
\textsc{ComparisonMul} is defined similarly. Finally, we define \textsc{PartWhole} as a simple set partition:
\begin{align*}
    &\forall v_t \forall X ( \\
    &\hspace{0.5em}\texttt{PartWhole}(X,v_t) \leftrightarrow \texttt{Partition}(X,v_t)).
\end{align*}
We also need a theorem indicating the the containers in a \textsc{PartWhole} relation have the same \texttt{Time}:
\begin{align*}
    &\forall v_t \forall X (\texttt{PartWhole}(X,v_t) \\
    &\hspace{0.5em}\to \forall v_s \in X (\texttt{Time}(v_s) = \texttt{Time}(v_t))).
\end{align*}

\subsection{Additional theorems for reasoning}

Additional theorems are required to enable reasoning about new containers from existing relations:
\begin{align*}
    &\forall e \forall r (\texttt{Transfer}(e) \land \texttt{Arg}(e,r) \\
    &\hspace{1.5em}\land \forall x\in r(\mathcal{E}(x) \land \mathcal{A}(x)) \\
    &\hspace{0.5em}\to \exists v \exists o \exists q (\texttt{Source}(e,v) \Land \texttt{Owner}(v,o) \\
    &\hspace{1.5em}\land \texttt{Measure}(v,q) \land \forall x \in v(\mathcal{E}(x) \land \mathcal{A}(x)).
\end{align*}
Note this is an axiom schema that is true for all entities $\mathcal{E}$ and attributes $\mathcal{A}$. We define the theorem similarly for the \texttt{Target} of $e$, as well as the case where the entities are uncountable.

We define similar theorems for the \textsc{Comparison} relation, omitted here for brevity. There is a minor difference in the theorem for the \textsc{Rate} relation, since its \texttt{Arg} is a set of sets:
\begin{align*}
    &\forall e \forall r (\texttt{Rate}(e) \land \texttt{Arg}(e,r) \\
    &\hspace{1.5em}\land \forall x\in r.\forall y \in x(\mathcal{E}(y) \land \mathcal{A}(y)) \\
    &\hspace{0.5em}\to \exists v \exists o \exists q (\texttt{Source}(e,v) \Land \texttt{Owner}(v,o) \\
    &\hspace{1.5em}\land \texttt{Measure}(v,q) \land \forall x \in v(\mathcal{E}(x) \land \mathcal{A}(x)).
\end{align*}

\section{Annotation Details} 
\label{sec:app-annotation}

\subsection{Data preprocessing} 
We segment all sentences into smaller independent clauses when possible. This is done in order to create simpler units of training data for a semantic parser. We use the Berkeley Neural Parser \citep{kitaev-klein-2018-constituency, kitaev-etal-2019-multilingual} for this task, splitting sentences recursively at the two coordinating conjunctions \emph{and} and \emph{but}.\footnote{Some phrases with a trailing preposition are split erroneously in this way, like ``Sally picked 7 lemons and Mary picked 9 lemons from the lemon tree'' is split into ``Sally picked 7 lemons'' and ``Mary picked 9 lemons from the lemon tree'', pointing to the challenges of prepositional phrase attachment in neural constituency parsing \citep{sopena-etal-1998-connectionist-approach}. We detect and correct such cases manually.} Over the three datasets we consider, $302$ sentences are split in this way. Additionally, some question sentences start with a subordinate clause that introduces new information, like ``If Alice bought 3 more apples today, how many apples did she end up with?''. We split these into a declarative clause and an interrogative clause, and remove the leading subordinating conjunction.

\subsection{Annotation scheme and tool} \label{sec:app-tool}
\begin{figure*}
    \centering
    \includegraphics[width=\textwidth]{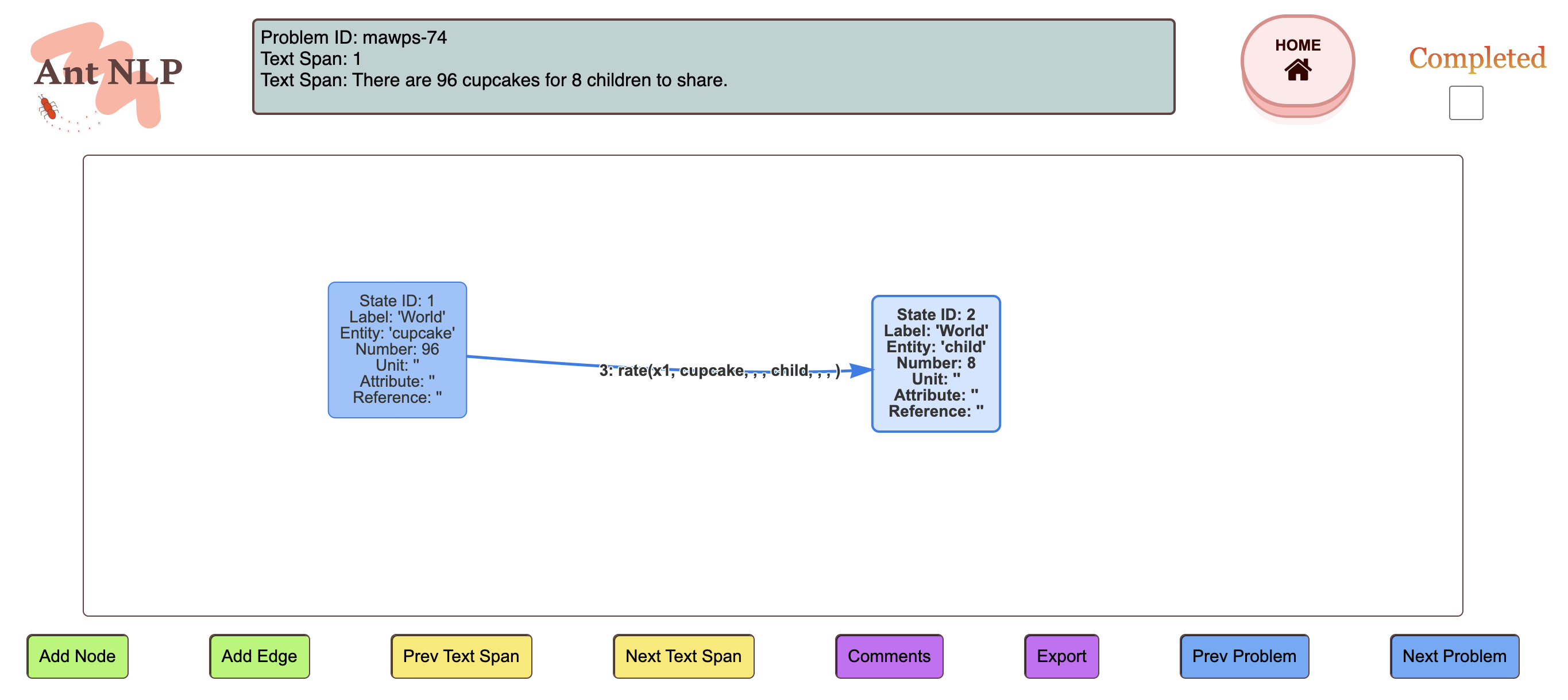}
    \caption{Snapshot of the annotation tool interface of \textsc{Ant-NLP} for a particular sentence. The annotator has the option to add nodes (containers), add edges (relations) between an existing pair of nodes, navigate between text spans, navigate between math story problems, add comments and flags associated with the math story problem, and export the math story problem when its annotation is completed. A problem is considered completed when all text spans have been annotated and a variable referencing the answer has been added for the final text span. Only then can the problem be marked in the top-right corner and be exported to json format. Furthermore, a central dashboard (not shown here) allows the annotator to get an overview over the progress and navigate between problems. Question sentences are not shown in the central dashboard in order to alleviate bias.}
    \label{fig:annotation-tool}
\end{figure*}

As mentioned in \cref{sec:world-model}, $\formalismname$ considers logical forms at the sentence level. Hence, we must also annotate the world model graphs incrementally, sentence by sentence. This is done via a drag-and-drop annotation tool, \textsc{Ant-NLP}, built specifically for the purpose of this work.\footnote{Although the annotation tool is built specifically for annotating world models in \formalismname, we believe it could with relative ease be adapted to annotation of potential world models for other domains as well. The tool will be shared with other researchers by request.} When annotating a problem in the tool, annotators get to build the graph incrementally one sentence at a time. Each sentence is given in a separate page, as shown in \cref{fig:annotation-tool}, and the graph from the previous sentence is carried over to the next. We save all incremental world models, as they set the basis for the sentence linearization described in \cref{sec:app-linearization}. The incremental world models are stored in json graph format.\footnote{\href{https://jsongraphformat.info/}{https://jsongraphformat.info/}}

We want annotators to include all information included in the text that fits $\formalismname$, irrespective of the relevance to the question. Therefore, in order not to create any bias stemming from the question, we hide the question sentence until all other preceding parts have been annotated. 

We ask annotators to follow the ordering that information is given within each sentence when adding containers and relations. For instance, a sentence ``Alice has 3 apples and 4 oranges'' should first be assigned a container for apples and then be assigned a container for oranges. This allows us to preserve the ordering of the text when linearizing logical forms for training data. To capture this ordering we annotate IDs for the containers and relations. The space of IDs is the set of natural numbers, and is shared between containers and relations. Ids are incremented automatically in the tool as annotators add new containers or relations.

The tool includes options to flag problems that require background knowledge, or where the annotator is uncertain about their annotation. They can additionally add a free text comment about their annotation for a particular problem.

\subsection{Procedure} 
Annotation is performed by external workers, who are taught to be familiar with the semantics of $\formalismname$. We employ two annotators, hired from a small annotation company in India.\footnote{There was initially a third annotator involved. However, this annotator dropped out during phase 3 as described below. At that time, it would have required a considerable time investment to hire and train yet another annotator, and so instead, we had one of the two other annotators cover up.} At the time of annotation, both of them were undergraduate students in technical universities. As support material, annotators are given a comprehensive guideline document, a set of example annotations and a video showcasing how annotation is performed using the tool. We follow three phases for annotation: 
\begin{enumerate}
    \item \textbf{Training phase}. This phase is for annotators to learn the formalism. They are given batches of $5-7$ problems at a time to annotate independently. These annotations are then discussed and, if needed, corrected in a follow-up meeting. The initial batches consist of simple problems, both conceptually and linguistically. After the annotators can successfully annotate these simple problems, they are gradually given more challenging ones. This phase ends when all annotators can successfully annotate most problems across all datasets.
    \item \textbf{Agreement phase}. Here, annotators are given the same set of $90$ problems, with $30$ from each dataset. They are asked to annotate these independently. This set is used to measure agreement between annotators.  
    \item \textbf{Scale-up phase}. Here, annotators are given separate datasets to annotate on their own. Some of these problems are overlapping in order to allow for agreement analysis.
\end{enumerate}

\subsection{Agreement analysis}
We give further details on the agreement analysis of the $125$ overlapping problems discussed in \cref{sec:data}. As mentioned, there were $18$ isomorphic disagreements between the two annotators (i.e., not weakly equivalent). Out of these, $7$ were due to structural ambiguity (\cref{sec:structural-ambiguity}), $1$ was due to a type of error that was fixed during annotation check (see below), $8$ due to a type of error for which a problem would be discarded during annotation check, and $2$ less serious errors that would not be detected during the annotation check. Most errors were attributed to the same annotator. Ground truth data for overlapping annotations were thus taken from the annotated set of the annotator with the higher performance on the overlapping problems.

There were $46$ problems that had a weak equivalence agreement, but not a strong equivalence agreement. Some of these were due to errors and some were due to property ambiguity (\cref{sec:property-ambiguity}). The errors were mostly incurred from entering an incorrect property, seemingly by carelessness. Several such cases could be detected and corrected as they led to errors when parsing the world model json file or when applying the reasoner to the world model (\cref{sec:app-reasoner}). Cases of property ambiguity were often due to the attribute property. 

We additionally stratified agreement across relation type. Problems with \comparison relations seemed to have the lowest weak equivalence agreement, followed by \textsc{Rate} and \textsc{Transfer}. For strong equivalence agreement on the other hand, \textsc{Rate} problems had the lowest agreement, followed by \textsc{Transfer} and then \textsc{PartWhole}.

\subsection{Annotation check and correction}
\label{sec:app-annotation-errors}
We performed the following checks of the annotations: whether the json could be parsed into a well-formed world model, whether applying the deterministic reasoner (\cref{sec:app-reasoner}) would produce the correct answer and whether the annotator had flagged the problem with low confidence or provided a free text comment. Based on these we were able to detect and correct several faulty annotations. Some common errors were: entering the wrong number, entering the wrong reference variable, forgetting to the enter the reference variable, orienting the edge in the wrong direction and misspelling label names. Such errors could easily be corrected. Other more fundamental errors that could not be easily fixed led to discarding the annotation. We also spotted some cases of wrong annotated answers stemming from the original dataset, which were corrected.

\section{Conversion between World Model Graph and Logical Form}
\label{app:conversion}
As mentioned in \cref{sec:solver_framework}, an integral part of our proposed \formalismname solver framework, and working with \formalismname more generally as in \cref{sec:applications}, is the conversion between world models $\aworldmodel$ and logical forms $\lform{}$. In this appendix section we provide details of both directions of this conversion. Both directions of the conversion are lossy to some small degree, as is mentioned in \cref{fn:lossy-partwhole}.
\subsection{World model graph to logical form}
\label{sec:app-linearization}

Each logical form can be viewed as a incremental graph update that consists of containers and relations based on a sentence in the problem text, which is represented as a text sequence.

Containers and relations have varying arity, depending on which properties are present. This opens two possibilities. We may either split them into forms for each set of properties and have the property names explicit in the signatures (e.g., containers would have one representation each with arity $3$ and $5$, and two representations with arity $4$), or keep the property ordering consistent and give a default null token for missing properties. We opt for the latter, and set the default null token to be \texttt{none}.

We define the following predicates:
\begin{itemize}
    \item \texttt{container(label, quantity, entity, attribute, unit)}
    \item \texttt{transfer(recipient label, sender label, quantity, entity, attribute, unit)}
    \item \texttt{rate(label, quantity, source entity, source attribute, source unit, target entity, target attribute, target unit)}
    \item \texttt{difference(target label, source label, quantity, target entity, target attribute, target unit, source entity, source attribute, source unit)}
    \item \texttt{explicit(target label, source label, quantity, target entity, target attribute, target unit, source entity, source attribute, source unit)}
    \item \texttt{part(whole label, whole entity, whole attribute, whole unit, part1 label, part1 entity, part1 attribute, part1 unit, \dots, partn label, partn entity, partn attribute, partn unit)}
\end{itemize}

Note that for \explicit, the ``type'' property is lifted out and its value replaces ``comparison'' as the name of the predicates. We replace ``add'' and ``times'' by ``difference'' and ``explicit'', respectively, for practical reasons: We do not want the name of the operator that might be required to solve the problem to be confounded with the name of the predicate. Further note that the above predicates 
are overloaded in comparison to the ones mentioned in \cref{sec:world-model}. The reason for that is that we require additional information in order to match the linearization to existing incremental graphs (the other direction of the conversion, described in \cref{sec:consistency-checker}). For instance, consider two disconnected containers in a world model graph. If one wished to present them as connected with \rate, it would be sufficient to provide the quantity property to the rate. See, e.g., how in \cref{fig:overview}, quantity is provided as the only property in the \rate relation. The other properties given above would be redundant as they are already given in the containers. For a model to be able to orient that rate, however, it needs the additional information to match to the two existing containers.

Note that in the case of \transfer, there may be two associated edges in the graph if the properties ``recipient label'' and ``sender label'' both take values other than $\texttt{none}$. However, these are both represented by a single \texttt{transfer} predicate as above. \partwhole is the only relation whose arity varies, reflecting the number of subsets present in the \partwhole construction. An alternative would have been to have one predicate per edge, but that would have introduced redundancy.\footnote{However, a drawback of our \partwhole representation is that it assumes that all the part-whole edges are always introduced together in the same sentence. While this is mostly the case for the data we observe, we found the following exception: ``Next on her list are the homeless people where she spent a total of \$900.00. She gave \$325.00 to the first set of homeless families and \$260.00 to the second set of families. How much did she give to the last set of homeless families?''. This is one example showing that the conversion is slightly lossy.\label{fn:lossy-partwhole}}

A sentence-level logical form often contains multiple components of the above. In these cases, we follow the ordering as introduced in text, in line with the annotated IDs. If a relation is added together with its source and/or target containers, then the containers must always precede the relation in the ordering. We enforce that the source container always precedes the target container.

As an example, the logical form of the sentence ``In a friend group there are 5 football players and 3 tennis players'' is: 
\begin{align*}
\allowbreak
    & \texttt{container(friend group, 5, player,} \\ & \texttt{football, none)} \\ 
    & \texttt{container(friend group, 3, player,} \\ & \texttt{tennis, none)}
\end{align*}

Finally, a world model graph may have containers that have not been explicitly introduced in text. For instance, the two sentences ``Alice has 5 apples. She ate 2 of them.'' will be represented by a world model with two containers and a \transfer edge, but only the source container is explicitly mentioned in text (in the first sentence). When writing the world model graph as a logical form, we therefore discard the target container in this case. In general, this is done by discarding all containers that do not hold an explicit quantity, unless the sentence is interrogative. For interrogative sentences we want the logical form to represent the reference variable.

\subsection{Logical form to world model graph}
\label{sec:consistency-checker}

We now consider the other direction, namely that we have a sequence of logical forms $\lform{1}, \dots, \lform{n}$ on the form described in the previous section and wish to convert them to a world model graph $\aworldmodel$. 

For $\lform{1}$, we can trivially convert the logical form to a graph. Note that the relation predicates specify the properties needed to match the relation to containers as well, so if there is a relation predicate in the logical form but no source and/or target container, we can simply create those. For subsequent logical forms, we match the logical form to the graph created from preceding sentences. For relations, we must make sure that we do not create new containers linked by that relation, if any or both of those containers are already existing in the world model. We thus first match the properties corresponding to the source and target containers in the relation predicate to any possibly existing containers, and only create new ones if none are found. 
In addition, some sentences will just supply an update of an unknown quantity to a known value. In these cases, we do not create a new container, but match the quantity to one already existing so that we can preserve the structural information of that container.
We remark that in case that the matching with already existing containers in the world model returns multiple options, we default to the most recently created one. This turns out to work well for most cases, but could be one source of loss.

The reference variable corresponds to the logical form of the last sentence: Interrogative sentences are mapped to logical forms the same way as declarative sentences, and the reference variable is taken as the variable in the container or relation that matches the question's logical form.

Finally, for predicted logical forms, we first check the logical forms for syntactic well-formedness, keeping only the parts of the logical form that are well-formed. An additional (weak) check for semantic well-formedness may match the properties to the vocabulary of the \msp, along with special tokens like ``none'', ``world'' etc.

\section{Difficult Cases to Parse}
\label{sec:appendix-difficult-cases}

We estimate a high coverage of our formalism among \msps. However, although a problem might be within semantic and conceptual coverage of \formalismname, the text itself might prove challenging for a parser to interpret. Here, we present two problems that are captured by $\formalismname$ that put a high burden on the parser.

\noindent First, consider the following problem:

\vspace{0.6em}
\begin{minipage}{19em}
  The teacher distributes 4 candies to 2 students. Student A now has 2 more candies than Student B. Both students had 0 candies to begin with. How many candies does Student A have?
\end{minipage}
\vspace{0.3em}

\begin{figure}[h]
    \centering
    \includegraphics[width=0.45\textwidth]{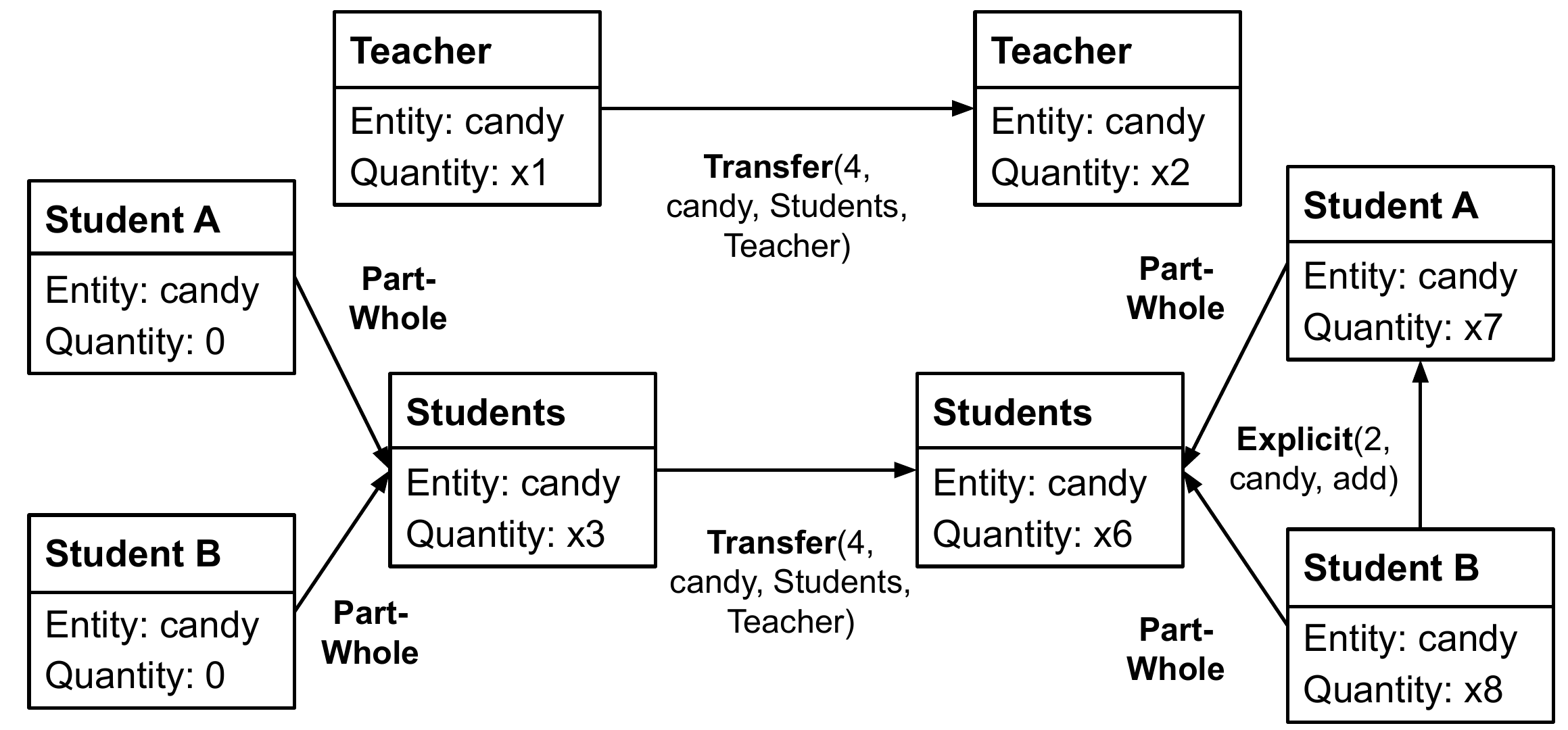}
    \caption{Hypothetical world model associated with a problem text that describes a subset-superset relation over containers.}
    \label{fig:world-model-students-candy}
\end{figure}

In this problem there is a transfer involved in the first sentence. The recipient of the transfer is not a single independent container however, but a set of two students. We have no information on how many candies these two students have individually, but we know that they collectively got $4$ more than they had before. To capture this, we may represent both students as a container with a \textsc{PartWhole} relation to the individual students, which will be the recipient of the \textsc{Transfer}. The whole problem is assigned the world model in \cref{fig:world-model-students-candy}. This is a faithful and correct world model, but the first sentence puts a high burden on the semantic parser: It must add $8$ containers and $6$ relations.

Next, consider the following problem (adapted from \textsc{GSM8K}):

\vspace{0.6em}
\begin{minipage}{19em}
    Zack decided to give his 3 friends 20 marbles each and kept 5. How many marbles did he initially have? 
\end{minipage}
\vspace{0.3em}

\begin{figure}[h]
    \centering
    \includegraphics[width=0.45\textwidth]{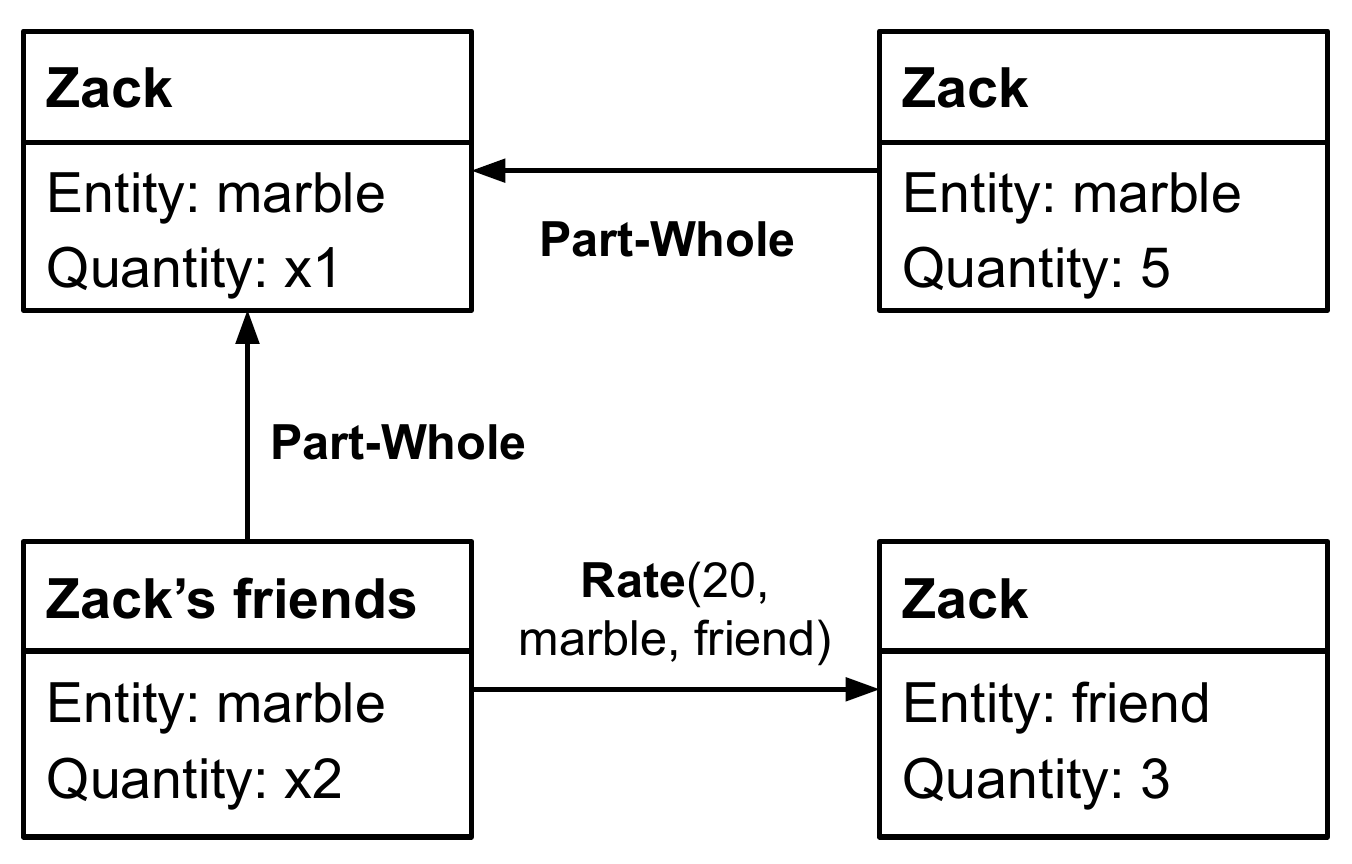}
    \caption{Hypothetical world model associated with a problem text that compresses a lot of information within a single sentence.}
    \label{fig:world-model-zack}
\end{figure}

The first sentence conveys a lot of information. We must add a container for the total number of marbles that Zack possesses, with \textsc{PartWhole} relations representing how many marbles Zack has left and how many his friends have. In addition, we know that there are three friends, which we represent with a \textsc{Rate}. See the world model in \cref{fig:world-model-zack}. However, the fact that Zack already possesses marbles is implicit from the text, and would be challenging for a parser to detect. As a partial remedy, we could introduce a ``TransferEvenly'' relation, which would represent a transfer of $20$ to each container in a set. In this case, Zack's friends would each be represented in a container.

\section{Reasoner}
\label{sec:app-reasoner}

The recursive solver takes as input a target variable and a set of visited equations. It takes all the equations containing the target variable and sorts them in increasing order of number of unknowns. Next, it iterates over the equations in this order. If the equation only has one unknown, that unknown must be the target variable. The function then solves for the target variable and outputs the numeric value. Otherwise, it goes over the other free variables in the equation and applies the recursive function to those as target, with the equation added to the set of visited equations in order to prevent loops. Having solved for the other free variable, it substitutes its numeric value in the equation and solves for the target variable, if possible. We present pseudo-code for the deterministic reasoner in \cref{algo:reasoner}.

Note that this solver assumes a certain structure of the equations, namely, that a solution can be reached by solving a sequence of equations with one unknown. Such is indeed the case for the simple \msps we consider. However, in the case of a general system of linear equations, this algorithm would fail as it cannot handle equations of more than one unknown. We opt for our recursive solution rather than Gaussian elimination due to runtime gains: for a system of $n$ equations with $n$ unknowns, Gaussian elimination runs in $\mathcal O(n^3)$, while our solution has worst-case complexity $\mathcal O(n^2)$.

Further note that if we extend $\refexpr$ to be a set of variables, we can store the intermediate results in a table and get a dynamic program. This is not necessary in our case as we do not have overlapping sub-problems.

\begin{algorithm}[ht!]
\footnotesize
\let\oldnl\nl
\newcommand{\nonl}{\renewcommand{\nl}{\let\nl\oldnl}}
\SetNlSty{}{\color{red}\sffamily}{}
\SetAlgoBlockMarkers{}{}
\SetKwProg{Fn}{function}{}{}
\SetKwIF{If}{ElseIf}{Else}{if}{ }{else if}{else }{}
\SetKw{Continue}{continue}
\SetKwFunction{FRecursiveReasoner}{\small recursiveReasoner}
\SetKwFor{For}{for}{}{end}
\SetKwProg{uForEach}{for each}{ do}{}
\SetKwProg{Fn}{function}{}{}
\AlgoDisplayBlockMarkers\SetAlgoVlined
\SetAlCapNameFnt{\small}
\SetAlCapFnt{\small}
\SetNoFillComment
\DontPrintSemicolon
\SetInd{-0.07em}{0.7em}
    \Fn{\FRecursiveReasoner{x, \texttt{visited}}}{
        \tcc{Prepare equations containing $x$ that have not been visited}
        $\texttt{eqs} \gets \{\text{equations containing }x\} \setminus \texttt{visited}$ \;
        \tcc{Sort in increasing order of number of unknowns}
        $\texttt{eqs} \gets \textbf{sort}(\texttt{eqs}, \# \text{ of unknowns}, \text{increasing})$ \;
        \For{$\texttt{eq} \in \texttt{eqs}$}{ 
            \tcc{Go over equations in order}
            \uIf{$\textbf{solvable}(x, \texttt{eq})$}{
                \tcc{Solve for $x$ if possible}
                $x_{\text{val}} \gets \textbf{solve}(x, \texttt{eq})$ \;
            }\Else(\tcc*[f]{Otherwise, solve recursively}){
                \tcc{Go over other unknowns}
                \For{$x' \in \texttt{eq}, x' \neq x$}{
                    $x'_{\text{val}} \gets \textbf{recursiveReasoner}(x, \texttt{visited}+\texttt{eq})$ \;
                    \tcc{Substitute unknown for value}
                    $\texttt{eq} \gets \textbf{substitute}(\texttt{eq}, x', x'_{\text{val}})$ \;
                    \If{$\textbf{solvable}(x, \texttt{eq})$}{
                        $x_{\text{val}} \gets \textbf{solve}(x, \texttt{eq})$ \;
                        \Return{$x_{\text{val}}$}
                    }
                }
            }
        }
        \Return{$x_{\text{val}}$}
	}
    \caption{Deterministic recursive reasoner.}
    \label{algo:reasoner}
\end{algorithm}

\section{Experimental Details}

\subsection{Solving pipeline}\label{sec:msp_solving_details}

\paragraph{Setup.} 
As our \llm we use Codex code-davinci-002. We design a prompt with 50 ground truth examples from \mawps and \asdiv. One example consists of the source sentence, the target linearized logical form, as well as the source and target of the previous sentence in the same \msp, in order to allow the model to account for dependencies between sentences. These examples are handpicked to be representative of \formalismname. For every \msp, we then feed each sentence following the same pattern excluding the target as a suffix to the prompt, and sample the target output from Codex. The experiments were performed on the 18th of January 2023. The parameters used for sampling were the following: temperature is set to $0$, max tokens is $200$, frequency and presence penalty are both left at $0$ and we add an additional new line stop token (which is used in the prompt to end the ground truth logical forms.

World models are built incrementally using the method described in \cref{sec:consistency-checker}. We apply the deterministic reasoner (\cref{sec:app-reasoner}) to produce an answer.\looseness=-1

\paragraph{Results.} We show the results in \cref{table:codex}. 
Observe that on average less than half of the predicted world models result in an answer (i.e. are complete). The rest of the times the reasoner is either unable to solve for the reference variable (the system of equations induced by the world model is underdetermined) or the world model lacks a reference variable. Incorrect answers are often caused by slight permutations of the correct logical forms (e.g., Codex having swapped the sender and recipient in a \transfer relation).
If we stratify the problems by relation type, we observe that the model has the highest answer accuracy for \transfer and \rate, while \partwhole problems have the lowest answer accuracy. This is to be expected given that the information associated with \partwhole problem is not often made explicit in text (\cref{sec:part-whole}).

\begin{table}[]
\fontsize{10}{10}\selectfont
\centering
\renewcommand{\arraystretch}{1.35} 
\setlength{\tabcolsep}{0.2em} 
\begin{tabular}{l|ccc}
    & \textsc{MAWPS}   & \textsc{ASDiv-A}      & \textsc{SVAMP}            \\ \hline
Answer Acc (\%) & 33.8    & 26.9   & 11.1            \\
Complete WM (\%) & 50.7  &  43.3  &  33.3 \\ 
Weak Smatch (avg.) & 0.76    & 0.68   & 0.59           \\
Strong Smatch (avg.) & 0.76    & 0.60   & 0.38          
\end{tabular}
\caption{Results on our test sets for the Codex few-shot learning model. Smatch scores (\cref{sec:app-smatch}) are averaged across all \msps, including those where Codex produced an incomplete world model.}
\label{table:codex}
\end{table}

\subsection{Details on constrained generation}\label{sec:generation_details}
The GPT 3.5 Turbo generation experiments were performed on the 24th of May 2023. The model used was gpt-3.5-turbo-0301. The sampling parameters are the same as those used during parsing (\cref{sec:msp_solving_details}). 

We display the results of the other five \msps as mentioned in \cref{sec:generation} in \cref{table:constrained_generation}. Observe that in all cases, the model is able to generate problems that are faithful to the concept, number and properties of the original world model (comparing the left column and the middle column). Further note that with a temperature parameter of $0$, the generated problems are rather conservative. We leave for future work to explore the implications of the sampling parameters for the generated outputs. Finally, consider the right column, where we display the \msps generated from augmented world models. Three of the generated examples are not completely faithful to how we augmented the world models. In the first example from the top, ``Lexie's brother'' is provided as the recipient property in the \transfer relation, but in the generated example Lexie's brother is the sender. In the third example from the top, we augment the world model with a \rate, but the model instead generates a transfer type \msp. In the last example, Bob is provided as sender while Josh is provided as recipient, but the model generates a problem with these values being swapped. The other two are faithful. 

\renewcommand{\cellalign}{tl}
\renewcommand{\theadalign}{tl}
\begin{table*}[t]
\tiny
\centering
\renewcommand{\arraystretch}{1.35} 
\setlength{\tabcolsep}{0.4em} 
\begin{tabular}{l|l|l|l}
\textbf{Original MSP}     &  \textbf{MSP generated from world model} & \textbf{Augmentation} & \textbf{MSP generated from augmented world model}                \\ \toprule
\makecell{Lexie's younger brother helped pick up \\ all the paper clips in Lexie's room. \\ He was able to collect 81 paper clips. \\ If he wants to distribute the paper clips in 9 boxes, \\ how many paper clips will each box contain?}       & \makecell{Lexie's brother has 81 paper clips.\\ He wants to put them in 9 boxes.\\ How many paper clips will he put in each box?} &  \makecell{\rate \\$\rightarrow$ \transfer}
& \makecell{Lexie's brother had 81 paper clips. \\ {\color{red} He gave 21 to Lexie.} \\ How many paper clips does Lexie's brother have now?}    \\ \hline

\makecell{Kevin collected toys to use as prizes at the fair. \\ He collected 14 stuffed animals. \\He also collected 18 frisbees and several yo-yos. \\ Kevin has 50 prizes in all. \\ How many yo-yos did Kevin collect?
} & \makecell{Kevin won 50 prizes at the fair. \\ He won 14 stuffed animals, 18 frisbees,\\ and some yo-yos. \\ How many yo-yos did he win?} & \makecell{\partwhole \\$\rightarrow$ \transfer}
& \makecell{ Kevin has 14 stuffed animals. \\ He gave 5 of them to his friend. \\ How many stuffed animals does Kevin have now?
} \\\hline

\makecell{Mrs. Hilt wants to make a border around her garden. \\ She needs 125 rocks to complete the border. \\ She has 64 rocks. \\ How many more rocks does she need to \\ complete the border?
} & \makecell{ Mrs. Hilt was making a rock border around her garden.\\ She had 125 rocks to use.\\ She used 64 rocks to make the border.\\ How many rocks did she have left?
} & \makecell{\transfer \\$\rightarrow$ \rate} & \makecell{ Mrs. Hilt is making a rock border around her garden.\\ She has 125 rocks to use.\\ {\color{red} She has already used 25 rocks.}\\ {\color{red} How many rocks does she have left?}
} \\\hline

\makecell{Zoe's school sold 620 dollars in raffle tickets. \\ If each ticket cost 4 dollars, \\ how many tickets did they sell?
} & \makecell{Zoe spent \$620 on raffle tickets.\\ Each ticket cost \$4.\\ How many tickets did she buy?} & \makecell{\rate \\$\rightarrow$ \transfer} & \makecell{Zoe had 620 dollars.\\ She spent 100 dollars.\\ How much money does Zoe have now?
} \\\hline

\makecell{Josh had 16 marbles in his collection. \\ He lost 7 marbles. \\ How many marbles does he have now?} & \makecell{Josh had 16 marbles.\\ He lost 7 of them.\\ How many marbles does Josh have now?} & \makecell{\transfer \\$\rightarrow$ \transfer} & \makecell{Josh has 16 marbles.\\ {\color{red} He gave 7 marbles to Bob.}\\ How many marbles does Josh have now?
} \\\bottomrule

\end{tabular}
\caption{Example of generated math story problems conditioned on world models in \formalismname. The left column shows the original math story problem, the middle column shows a math story problem generated conditioned on the ground truth world model of the original problem, and the right column shows a math story problem generated conditioned on a world model that has been created by augmenting the ground truth world model of the original problem. Sentences not faithful to the logical form are colored {\color{red} red}.}
\label{table:constrained_generation}
\end{table*}

\subsection{Details on prompting using synthetic questions}
\label{sec:synthetic_prompt_templates}

\begin{table*}[t]
\fontsize{11}{11}\selectfont
\centering
\renewcommand{\arraystretch}{1.35} 
\setlength{\tabcolsep}{0.4em} 
\begin{tabular}{l|l}
containers     & Q: How many \{attr\}\{ent\}s does \{label\} have? A: \{quant\}                    \\
               & Q: What is the amount of \{attr\}\{ent\}s associated with \{label\}? A: \{quant\} \\ \hline
\transfer       & Q: How many \{ent\}s are transferred from \{sour\} to \{targ\}? A: \{quant\}             \\
\explicit (add)   & Q: How many more \{ent\}s does \{targ\} have than \{sour\}? A: \{quant\}          \\
\explicit (times) & Q: How much more \{ent\} does \{sour\} have than \{targ\}? A: \{quant\}           \\
\rate           & Q: How many \{ent\} does \{targ\} have per \{sour\}? A: \{quant\}                 \\
\partwhole     & Q: How many \{sour\} are part of \{targ\}? A: \{quant\}                          
\end{tabular}
\caption{Templates to automatically create question-answer pairs for prompting. The templates are filled based on the information in the world model.}
\label{table:synthetic_prompt_templates}
\end{table*}

\begin{table*}[t]
\small
\centering
\renewcommand{\arraystretch}{1.35} 
\setlength{\tabcolsep}{0.4em} 
\begin{tabular}{l|l}
\textbf{prompt types} & \textbf{pairs sourced from one MSP (one-shot), i.e., $x = 1$} \\ \hline
\begin{tabular}[c]{@{}l@{}}(1) synth QAs \\ (all at once)\end{tabular} & \begin{tabular}[c]{@{}l@{}}Baker made 43 cakes and 114 pastries. If he sold 154 pastries and 78 cakes.\\ Q: How many cakes does baker have?\\ A: 43\\ Q: How many sold cakes are associated with baker?\\ A: 78\\ Q: How many more pastries than cakes did baker sell?\\ A: 76\\ \\ Bobby had 19 pieces of candy. He ate 2 pieces of candy.\\ Q: What is the amount of candys associated with bobby?\\ A: 19\\ Q: How many candys are transferred from bobby?\\ A: 2\\ Q: how many pieces of candy does he still have left?\\ A:\end{tabular}                                                                                    \\ \hline
\begin{tabular}[c]{@{}l@{}}(2) synth QAs\\ (sent by sent)\end{tabular} & \begin{tabular}[c]{@{}l@{}}Baker made 43 cakes and 114 pastries. \\ Q: How many cakes does baker have?\\ A: 43\\ Baker made 43 cakes and 114 pastries. If he sold 154 pastries and 78 cakes.\\ Q: How many sold cakes are associated with baker?\\ A: 78\\ Baker made 43 cakes and 114 pastries. If he sold 154 pastries and 78 cakes.\\ Q: How many more pastries than cakes did baker sell?\\ A: 76\\ \\ Bobby had 19 pieces of candy. \\ Q: What is the amount of candys associated with bobby?\\ A: 19\\ Bobby had 19 pieces of candy. He ate 2 pieces of candy. \\ Q: How many candys are transferred from bobby?\\ A: 2\\ Bobby had 19 pieces of candy. He ate 2 pieces of candy.\\ Q: how many pieces of candy does he still have left?\\ A:\end{tabular} \\ \hline
\begin{tabular}[c]{@{}l@{}}(3) original\\  \msp QAs\end{tabular}        & \begin{tabular}[c]{@{}l@{}}Baker made 43 cakes and 114 pastries. If he sold 154 pastries and 78 cakes.\\ Q: How many more pastries than cakes did baker sell?\\ A: 76\\ \\ Bobby had 19 pieces of candy. He ate 2 pieces of candy.\\ Q: how many pieces of candy does he still have left?\\ A:\end{tabular}                                                                 
\end{tabular}                                        
\caption{We experiment with three different types of prompts. They are displayed for the one-shot case in which one \msp in addition to the one we are trying to solve is provided in the prompt. In the above case, the model is tasked with making inference on the problem ``Baker made 43 cakes and 114 pastries. If he sold 154 pastries and 78 cakes. How many more pastries than cakes did baker sell?''.}
\label{table:prompt_types}
\end{table*}

The GPT-3 probing experiments were performed on the 18th of January 2023. The model used was text-davinci-003. The sampling parameters used are the same as those used for Codex during parsing (\cref{sec:msp_solving_details}).

In \cref{table:synthetic_prompt_templates}, we present the templates used to create synthetic question-answer pairs for prompting large language models.

\end{document}